\documentclass{article}

\usepackage{arxiv}

\usepackage[utf8]{inputenc} % allow utf-8 input
\usepackage[T1]{fontenc}    % use 8-bit T1 fonts
\usepackage{hyperref}       % hyperlinks
\usepackage{url}            % simple URL typesetting
\usepackage{booktabs}       % professional-quality tables
\usepackage{amsfonts}       % blackboard math symbols
\usepackage{nicefrac}       % compact symbols for 1/2, etc.
\usepackage{microtype}      % microtypography
\usepackage{algorithm}
\usepackage[table]{xcolor}
\usepackage{cite}
\usepackage{amsmath,amssymb,amsfonts}
\usepackage{algorithmic}
\usepackage{graphicx}
\usepackage{textcomp}
\usepackage[table]{xcolor}
\usepackage{booktabs, multirow}

\newcolumntype{L}{>{\centering\arraybackslash}m{0.11\linewidth}}
\newcolumntype{M}{>{\centering\arraybackslash}m{0.07\linewidth}}
\newcolumntype{N}{>{\centering\arraybackslash}m{0.09\linewidth}}
\newcolumntype{O}{>{\centering\arraybackslash}m{0.04\linewidth}}

\def\BibTeX{{\rm B\kern-.05em{\sc i\kern-.025em b}\kern-.08em
    T\kern-.1667em\lower.7ex\hbox{E}\kern-.125emX}}

\title{An "outside the box" solution for imbalanced data classification}

\author{
 Hubert Jegierski \\
  Faculty of Computer Science and Management\\
  Wrocław University of Science and Technology\\
  Wrocław, Poland \\
  \texttt{hubert.jegierski@gmail.com} \\
   \And
 Stanisław Saganowski \\
  Department of Computational Intelligence\\
  Faculty of Computer Science and Management\\
  Wrocław University of Science and Technology\\
  Wrocław, Poland \\
  \texttt{stanislaw.saganowski@pwr.edu.pl} \\
}

\begin{document}
\maketitle

\begin{abstract}
A common problem of the real-world data sets is the class imbalance, which can significantly affect the classification abilities of classifiers. Numerous methods have been proposed to cope with this problem; however, even state-of-the-art methods offer a limited improvement (if any) for data sets with critically under-represented minority classes. For such problematic cases, an "outside the box" solution is required. Therefore, we propose a novel technique, called enrichment, which uses the information (observations) from the external data set(s). We present three approaches to implement enrichment technique: (1) selecting observations randomly, (2) iteratively choosing observations that improve the classification result, (3) adding observations that help the classifier to determine the border between classes better.
We then thoroughly analyze developed solutions on ten real-world data sets to experimentally validate their usefulness. On average, our best approach improves the classification quality by 27\%, and in the best case, by outstanding 66\%. We also compare our technique with the universally applicable state-of-the-art methods. We find that our technique surpasses the existing methods performing, on average, 21\% better. The advantage is especially noticeable for the smallest data sets, for which existing methods failed, while our solutions achieved the best results. Additionally, our technique applies to both the multi-class and binary classification tasks. It can also be combined with other techniques dealing with the class imbalance problem.
\end{abstract}

% keywords can be removed

\section{Introduction}
\label{sec:introduction}
%intro for 'machine learning'
The Class Imbalance Problem (CIP) occurs when the distribution of data between considered classes is imbalanced, i.e., one of the classes, called the dominant or majority class, exceeds the size of the other classes (minority classes). The CIP is especially undesired in prediction/classification tasks as most of the existing algorithms will focus on the classification of the dominant class while ignoring or misclassifying under-represented observations. Unfortunately, against our will, real-world data sets tend to express an imbalanced nature. In result, the CIP is a well-known problem to many everyday situations like medical diagnosis prediction \cite{krawczyk2016evolutionary, yuan2018regularized, fotouhi2019comprehensive}, detecting frauds in banking operations \cite{soh2019predicting, shi2015novel, van2015apate}, image processing \cite{khan2017cost, feng2019imbalanced}, natural language processing \cite{wang2019research, al2017using, krawczyk2017sentiment}, real-time data streaming \cite{pouyanfar2018dynamic, hoens2012learning}, and other prediction/classification tasks \cite{cieslak2006combating, burez2009handling, pan2014graph, yan2015deep, kim2016ordinal}.

A great number of methods dealing with the CIP have been proposed. However, sometimes, they cannot be applied (a very often problem for small data sets), or their effectiveness is unsatisfactory. Therefore, we propose a new data set balancing technique called enrichment. Its main idea is adding observations from an external data set to the under-represented class(es) of the considered data set. Using ten real-world data sets we thoroughly analyze three different enrichment scenarios: (1) random selection of observations from external data set(s); (2) semi-greedy - iterative selection of observations from the external data set(s) which improve the final classification; and (3) supervised, in which only observations that help the classifier to determine the border between classes are selected from the external data set(s). We show that our solution outperforms other methods, especially for small data sets, where existing solutions were unable to improve the classification accuracy. On the considered data sets, we achieved, on average, 21\% better results than competitive techniques.

To emphasize, our contributions are: (1) a new technique, called data set enrichment, for dealing with the Class Imbalance Problem; (2) three approaches to implement data set enrichment; (3) the successful solution for balancing data sets with a critically low number of under-represented class (e.g. 4 observations); (4) the first-ever solution that improves the results of minority classes without affecting the result of the dominant class(es); (5) an experimental evaluation of the proposed solutions, including comparison with the state-of-the-art methods and the transfer learning technique.

\section{Related work}
\label{sec:related_work}

%Related work

Existing methods aiming to reduce the negative effect of the CIP can be divided into three types: (1) data-level methods focused on lowering the quantitative difference between classes, e.g., by sampling the majority class, or by multiplying examples of the minority class; (2) algorithm-level methods that modify or extend classifiers to deal with imbalanced classes, e.g., a classifier with the cost matrix - a misclassification of the minority class is much more expensive than a misclassification of the majority class; (3) hybrid methods combining the abilities of the two previous types.

The most often data-level concepts are sampling methods (the most popular), data cleaning techniques, and feature selection methods. The sampling technique can be performed as under-sampling, over-sampling, and mixed sampling. The under-sampling removes examples from the dominant class. It is effortless to perform but loses information when selecting a subset of the majority class. On the other hand, the reduced data set speeds up the learning process and reduces the required memory space. The simplest version of under-sampling is Random Under-sampling (RUS), explained in \cite{japkowicz2002class}, which randomly eliminates examples from the dominant classes. More advanced approaches include clusterization \cite{sun2015novel, yen2009cluster}, evolutionary algorithms \cite{jain2017addressing, krawczyk2016evolutionary}, instance hardness identification \cite{smith2014instance}, and weighting objects based on Euclidean distance \cite{kang2017distance}.

The over-sampling, on the contrary, increases the size of the minority set by duplicating existing observations or generating artificial ones. This approach is not losing information, but it requires a greater amount of computing power, and an inappropriate increase in the data set might lead to overfitting of the model. Again, the most simple approach, Random Over-sampling (ROS), explained in \cite{japkowicz2002class}, is based on multiplying randomly selected observations of the minority class. Another method, widely used, is Synthetic Minority Over-sampling Technique \cite{chawla2002smote} (SMOTE), which generates synthetic observations based on the selected observation and its neighbors. Many extensions and modifications to the SMOTE algorithm have been proposed, e.g., the Borderline SMOTE \cite{han2005borderline} (bSMOTE), which detects and uses the borderline observations to generate new synthetic samples. Other over-sampling concepts employ clusterization \cite{cieslak2006combating}, genetic algorithms \cite{li2017adaptive}, potential functions \cite{krawczyk2019radial}, and identification of harder to learn observations \cite{he2008adasyn}.

Mixed sampling is a combination of under-sampling and over-sampling. It combines benefits (and some limitations) of both approaches mentioned above. Some methods applying the mixed sampling technique are Selective Preprocessing of Imbalanced Data \cite{wojciechowski2017algorithm} (SPIDER), Optimal Genetic Algorithms based Resampling \cite{vannucci2017genetic} (OGAR), SMOTE-ENN \cite{batista2004study} combining SMOTE and Edited Nearest Neighbors \cite{wilson1972asymptotic}, and SMOTE-TL \cite{batista2003balancing} combining SMOTE and Tomek Links \cite{tomek1976two}.

The most significant advantage of sampling techniques is the possibility to apply them at the pre-processing stage, regardless of the further steps, i.e., classifiers used. This advantage allows for combining sampling techniques with algorithm-level methods for reducing the imbalanced data set effect. The most often applied algorithm-level technique is the cost-sensitive learning \cite{elkan2001foundations, domingos1999metacost}. The approach emphasizes the correct classification of the minority objects by assigning a high penalty for the misclassification of these objects. The penalty is stored in the form of a cost matrix. The advantage of cost-sensitive learning is its computational efficiency \cite{haixiang2017learning}. However, determining the values of the cost matrix is not trivial and might require many attempts. Additionally, this approach demands modification of the learning algorithm. Another popular technique is a classifier ensemble. Several classifiers perform classification of the same observations independently, and their output is later combined into an ensemble output. Usually, the most commonly assigned label is selected as the final output. The benefits of this approach include the ability to bypass the statistical problem, i.e., the learning set is too small, and the ability to reduce the bias of a single classifier. The shortcomings of the approach are the lack of clear guidelines for choosing the ensemble members and the lack of voting methods tailored to imbalanced data \cite{krawczyk2016challenges}. In \cite{sun2015novel}, each of the ensemble's classifiers is trained on a different, balanced training set, and in \cite{ksieniewicz2017dealing}, the visual representation of the numerical data distribution called \textit{exposer} is used.

There are also many algorithm-level strategies that extend existing learning algorithms, e.g., strengthen the discriminatory feature of classifiers \cite{raj2016towards}, use gravitational forces to determine the neighborhood in k nearest neighbor classifier \cite{nikpour2017proposing}, extend a classifier by fuzzy methods \cite{patel2017classification}. Yet another interesting approach is presented in \cite{krawczyk2014weighted}, where authors used one-class learning. By focusing on a single class, the method develops an accurate description of the target class, and thus classify it very well, rejecting outliers.

In contrast to the imbalanced binary classification, the multi-class classification from imbalanced data is not as well developed. A common approach is the decomposition of a complex problem to a set of binary tasks. The two most widely used techniques include one-vs-one (OVO), and one-vs-all (OVA) approaches. The OVO strategy divides the problem into subtasks, where each of them is aimed at distinguishing a given pair of classes. In turn, the OVA approach trains every classifier so that it recognizes a given class. However, such solutions have their drawbacks: OVA loses information on the relationship between classes, while OVO leads to excessive complexity related to the analysis of all possible combinations.

The last group of methods dealing with imbalanced data combines data-level and algorithm-level concepts. An example is an ensemble of classifiers enhanced with a sampling technique \cite{nanni2015coupling}.

% successful examples for 'complex networks'
%The existing methods dealing with imbalanced data provide quite satisfactory results when applied to the complex networks domain. The SMOTE method helped in predicting the tie strength in network \cite{sohrabi2016comprehensive} and in detection of intrusions into systems \cite{cieslak2006combating}. The RUS was applied to increase the accuracy of the churn prediction \cite{burez2009handling, rehman2015customer}. The sampling technique mixed with ensemble classifiers were utilized to boost link prediction in collaborative networks \cite{lakshmi2017temporal, lakshmi2017link}. Extending the random forest by extracting features improved detection of credit card frauds \cite{van2015apate}.

Although, there are many examples showing the usefulness of methods dealing with the CIP \cite{krawczyk2016evolutionary, yuan2018regularized, fotouhi2019comprehensive, soh2019predicting, shi2015novel, van2015apate, khan2017cost, feng2019imbalanced, wang2019research, al2017using, krawczyk2017sentiment, pouyanfar2018dynamic, hoens2012learning, cieslak2006combating, burez2009handling, pan2014graph, yan2015deep, kim2016ordinal}, there are cases when existing methods cannot be applied, or their performance is low. This limitation is especially noticeable for data sets with a low number of observations and data sets with a significant imbalance ratio. When there are only a few observations of a minority class, the richness of information is to low to produce diverse synthetic observations using the over-sampling technique. At the same time, under-sampling the majority class to only a few observations will result in a set too small to successfully train the classifier.

Therefore, in this paper, we propose a new technique called data set enrichment, which consists of adding observations of the minority class(es) from the external data set(s). We propose three enrichment scenarios: random, semi-greedy, and supervised. We show that our solution often outperforms other existing methods, especially for small data sets, where existing solutions were unable to improve the classification accuracy. On the considered data sets, we achieved, on average, 21\% better results than when using competitive techniques.

\section{Materials and Methods}
\label{sec:materials_&_methods}

Before explaining our methods, we first need to introduce the Imbalance Ratio formula and explain various types of imbalanced objects.

\subsection{Imbalance ratio}

The ratio between the dominant class and other classes can be expressed with the Imbalance Ratio ($IR$) value; see \ref{eq:1}. The minority classes can be then defined as classes with $IR$ below a certain threshold, e.g., 0.2.

%In order to reduce the degree of unbalancing, only classes that constitute a minority in the base set are used.

\begin{equation}
IR = \frac{|M_b|}{|D_b|},
\label{eq:1}
\end{equation}

where $D_b$ is a subset of data set $b$ that is labeled with the dominant class and $M_b$ is a subset of data set $b$ that is labeled with the considered, non-dominant class.

\subsection{Object types}

\begin{figure}
  \centering
  \includegraphics[width=\textwidth]{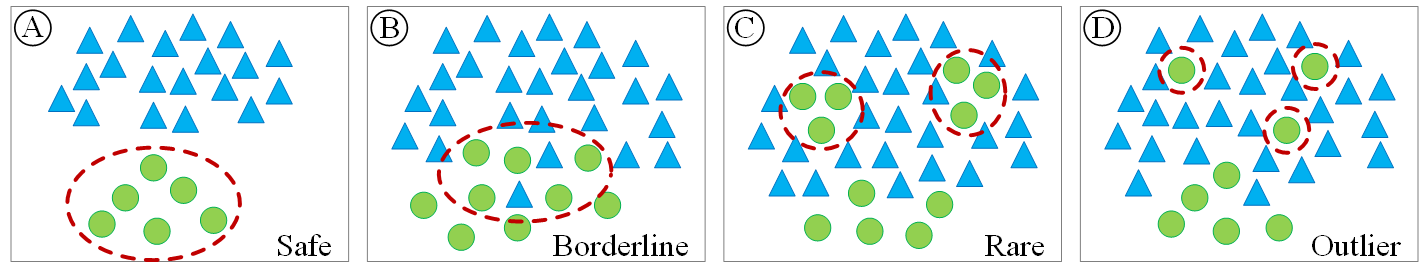}
  \caption{Four types of minority class objects: (\textbf{A}) safe objects, (\textbf{B}) borderline objects, (\textbf{C}) rare objects, (\textbf{D}) outlier objects.}
  \label{fig_observation_types}
\end{figure}

Napierała and Stefanowski, in \cite{napierala2016types}, focused on the neighborhood of minority class objects. They identified four scenarios for which they introduced four types of minority objects:
\begin{itemize}
\item    Safe objects (Fig. \ref{fig_observation_types}A) – which are neighboring only with objects of the same class, and are distanced from the potential class decision boundary. These examples are simple to learn by classifiers.
\item    Borderline objects (Fig. \ref{fig_observation_types}B) – objects located near the potential class decision boundary, important for determining the location of the separating plane of the class. These examples are very often misclassified as the dominant class.
\item    Rare objects (Fig. \ref{fig_observation_types}C) – objects clustered in small groups of several items, located inside the dominant class, away from the center of the associated minority class. These are examples that should be given special attention, as they tend to be ignored (considered as a noise) by a classifier, and they often constitute a large part of all minority observations \cite{stefanowski2013overlapping}.
\item    Outlier objects (Fig. \ref{fig_observation_types}D) – individual minority objects located inside the dominant class. Outliers should not be considered as noise, because they can create subgroups of valid objects, and inform about the potential area of the appearance of new minority objects.
\end{itemize}

We use their notation in our method.

\subsection{Random Enrichment}

The first of the proposed methods of enrichment is Random Enrichment (RanE), which consists of adding random learning patterns from the external data set(s) to the considered base set. Algorithm~\ref{alg1_ranE} provides the pseudo-code of this approach. At first, the minority classes $M_b$ of the base set $b$ are determined using $IR$ formula~\ref{eq:1}. Next, the corresponding classes $M_e$ in the external data set(s) $e$ are identified. Finally, a specified percentage amount $A\%$ of observations from classes $M_e$ are added to the base set $b$, resulting in the enriched data set $b^\prime$.

\renewcommand{\algorithmicrequire}{\textbf{Input:}}
\renewcommand{\algorithmicensure}{\textbf{Output:}}
\begin{algorithm}%\onehalfspacing
\caption{Random Enrichment}
\label{alg1_ranE}
\begin{algorithmic}
    \REQUIRE base set $b$, external set $e$, imbalance ratio $IR$ among classes in $b$, amount of enrichment $A\%$
    \ENSURE enriched set $b^\prime$
    \STATE determine minority classes $M_b$ in set $b$ based on $IR$
    \STATE identify corresponding classes $M_e$ in set $e$
    \STATE sample $A\%$ of set $M_e$ and add it to base set
    \RETURN $b\prime$
\end{algorithmic}
\end{algorithm}

\subsection{Semi-greedy Enrichment}

\begin{figure}
  \label{fig_seme_supe}
  \centering \includegraphics[width=\textwidth]{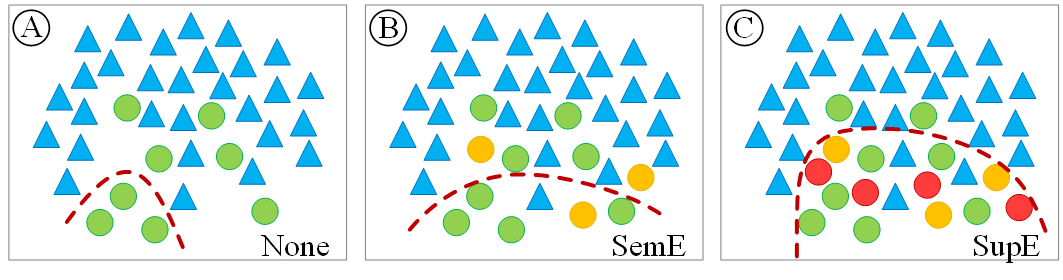}
    \caption{An example of how the base set (\textbf{A}) can be enriched with external observations using the SemE (\textbf{B}) and the SupE (\textbf{C}) methods. The external observations added with SemE and SupE are marked yellow and red, respectively.}
\end{figure}

The second proposed approach is called Semi-greedy Enrichment (SemE), as it iteratively validates, which observations from the external data set(s) will increase the classification performance. See Fig.~\ref{fig_seme_supe}B for an example and  Algorithm~\ref{alg2_semE} for the pseudo-code. The minority classes $M_b$ of the base set $b$, and the corresponding classes $M_e$ in the external data set(s) $e$ are recognized. The base set $b$ is divided into learning part $b_l$ and evaluation part $b_{ev}$ with ratio 9:1 using stratified sampling. Using the preferred classifier $C$ (in our case Random Forest), the baseline classification performance $C_{best}$ is established. Then, iteratively, one observation $O$ at a time from class(es) $M_e$ is added to the base learning set $b_l$ and the new classification performance $C_{new}$ is computed using $b_{ev}$. If performance has improved, observation $O$ is kept in set $b_l$, and the baseline classification performance is updated; otherwise, observation is discarded. The iterative process is repeated until all observations from $M_e$ are evaluated. The final output will be enriched set $b^\prime$, in which the classification performance will be at least as good as the classification performance of the base set $b$.

\begin{algorithm}%\onehalfspacing
\caption{Semi-greedy Enrichment}
\label{alg2_semE}
\begin{algorithmic}
    \REQUIRE base set $b$, external set $e$, imbalance ratio $IR$ among classes in $b$, classifier $C$
    \ENSURE enriched set $b^\prime$
    \STATE determine minority classes $M_b$ in set $b$ based on $IR$
    \STATE identify corresponding classes $M_e$ in set $e$
    \STATE divide set $b$ into learning part $b_l$ and evaluation part $b_{ev}$ with ratio 9:1 using stratified sampling
    \STATE calculate baseline classification performance $C_{best}$ using classifier $C$
    \FOR{ \textbf{each} observation $O$ \textbf{in} $M_e$}
        \STATE add $O$ to $b_l$ and evaluate classification performance $C_new$
        \IF{$C_{new} > C_{best}$}
            \STATE keep $O$ in $b_l$
            \STATE $C_{best} = C_{new}$
        \ELSE
            \STATE remove $O$ from $b_l$
        \ENDIF
    \ENDFOR
    \RETURN $b\prime$
\end{algorithmic}
\end{algorithm}

\subsection{Supervised Enrichment}

The last developed method, called Supervised Enrichment (SupE), assumes that only external observations of a particular type are added. In our study, we have only used the borderline observations, as these help classifiers in defining boundaries between classes, see Fig.~\ref{fig_seme_supe}C for an example. The pseudo-code of the algorithm is presented in Algorithm~\ref{alg3_supE}. At first, the minority classes $M_b$, and the corresponding classes $M_e$ are determined. Then, the SemE algorithm can be applied to increase the number of observations in classes $M_b$. This extra step will help in finding more borderline observations from the external data set. An alternative approach to SemE at this point could be an over-sampling technique within rare observations. Observations added during SemE step (if any) are removed from the external set $e$ to assure they will not be added again in further steps. Next, observations in $M_e$ are labeled with a type they will have when added to set $b$. The type is determined according to the method by Napierała and Stefanowski \cite{napierala2016types} (see the next section) with $k$ parameter equal to five. Finally, the borderline objects are added to set $b$ until $IR$ among classes $M_b$ and class $D_b$ equals one or until all available borderline observations are added. The final set is enriched set $b^\prime$.

The decision whether to use an additional step (SemE or an over-sampling technique) depends on the number of external observations that one will be able to obtain without this extra step. If the minority class is too small, it might be difficult (or even impossible) to find observations of the desired type.

%In the case where the number of borderline objects is too small, the neighborhood of the base set should be adjusted to the state in which borderline objects could occur. For this purpose, semigreedy enrichment or synthetic object generation within rare objects can be used. This paper uses semigreedy enrichment before the borderline objects search stage.

\begin{algorithm}%\onehalfspacing
\caption{Supervised Enrichment}
\label{alg3_supE}
\begin{algorithmic}
    \REQUIRE base set $b$, external set $e$, imbalance ratio $IR$ among classes in $b$, classifier $C$
    \ENSURE enriched set $b^\prime$
    \STATE perform SemE algorithm
    \STATE remove from $e$ observations added to $b$ with SemE
    \STATE obtain borderline observations from $M_e$
    \STATE add obtained observations to $b$ until $IR = 1$ or all observations are added
    \RETURN $b\prime$
\end{algorithmic}
\end{algorithm}

\section{Experiments}

One of our goals was to enrich the base set with as many different external data sets as possible. This would allow evaluating if (1) the considered base set prefers a particular external data set, and (2) a particular external data set can be more useful than others to several base sets. To achieve this goal, we required several different data sets with the same multiple classes labeled, and with the same feature space. Hence, we have used eight real-world complex network data from our previous extensive work on group evolution prediction \cite{saganowski2019gep} that are publicly available \cite{saganowski2018data}. In addition, we have used Breast Tumor (BRE) \cite{dataset_BRE} and Prostate Tumor (PRO) \cite{dataset_PRO} data to validate the solution for the binary classification task and also outside the context of the complex networks. Each data set has a different size and the class imbalance ratio, see Appendix \ref{appendix:results_by_class}. For the classification task on the complex network data sets, six classes are determined by 88 features. For the BRE and PRO data sets, two classes are described by 30 and 8 features, respectively. The F-measure averaged over all classes is used as the quality performance measure. The detailed information on data used, pre-processing, and on the experimental setup is provided in Appendix~\ref{appendix:experiment_setup}.

\subsection{Random Enrichment}

The effectiveness of Random Enrichment (RanE) is presented in Table \ref{tab_ranE}. It can be observed that it is possible to find a random subset of an external data set that improves the classification quality of the considered base set $b$ (in this case, the MIT data set). Let us emphasize that for each external data set, such a subset has been found (cells marked with green color in Table \ref{tab_ranE}). It can also happen that randomly sampled observations from the external data set(s) added to the base set results in lower classification quality (cells marked red). For some external data sets (in this case, Digg, IrvineMessages, Slashdot, Infectious), all randomly selected subsets improved the classification quality, and for other (Loans, Facebook, Twitter) only a few subsets boosted the classification. The best scenario, adding 90\% of the Infectious $M_e$ classes, improved the classification of the MIT data set by +20\% (cell marked with yellow color in Table~\ref{tab_ranE}). The Infectious was the best external data set to enrich the MIT data set, on average, boosting the classification by 12\%.

The other 7 data sets related to complex networks have also been successfully randomly enriched. The full results are provided in Appendix~\ref{appendix:rane_full_results}.

Despite many analyses, we can not provide any formal rules on how to select the enrichment set to obtain a better classification on the base set with the RanE method. No correlation was found with either the number of observations added nor the characteristics of the external or base data sets (size, network degree, network betweenness, and other network measures). Most probably, the quality (the type as defined by Napierała and Stefanowski) of observations being added is the crucial factor behind the successful random enrichment. In that sense, the procedure of the algorithm, i.e., the random selection of a subset, determines the outcome. Hence, the name of the method - Random Enrichment.

\begin{table}[H]
\caption{The performance of the RanE method for the MIT data set with different external data sets and various amounts of external observations. The method improved the classification quality by up to 20\%. The classification executed with the Decision Tree classifier and the F-measure averaged over all classes as the quality measure. Cells marked green indicate an improved classification quality, while cells marked red depict worse results. The best result of each external data set is bolded, and the best overall result is marked yellow.}
\label{tab_ranE}
\centering
\setlength\tabcolsep{2pt}
\begin{tabular}{c|cLccccc}
\toprule
\multirow{2}{1.5cm}{\centering\vfil\centering\bfseries\textbf{Amount of $M_e$}}    & \multicolumn{7}{c}{\bfseries External data set}\\ \cmidrule(lr){2-8}   & \textbf{Digg} & \textbf{Irvine Messages}   & \textbf{Loans}    & \textbf{Facebook}   & \textbf{Slashdot} & \textbf{Infectious} & \textbf{Twitter}\\
\midrule
0\%	& \multicolumn{7}{c}{0.210}\\
10\%	& \cellcolor{green!25}0.219 +4\%	& \cellcolor{green!25}0.222 +6\%	& \cellcolor{red!25}0.209 -0\%	& \cellcolor{red!25}0.208 -1\%	& \cellcolor{green!25}0.218 +4\%	& \cellcolor{green!25}0.233 +11\% & \cellcolor{red!25}0.192 -9\%\\
20\%	& \cellcolor{green!25}\textbf{0.233 +11\%}	& \cellcolor{green!25}0.222 +6\%	& \cellcolor{green!25}0.217 +3\%	& \cellcolor{green!25}0.213 +1\%	& \cellcolor{green!25}0.213 +1\%	& \cellcolor{green!25}0.215 +2\%   & \cellcolor{red!25}0.197 -6\%\\
30\%	& \cellcolor{green!25}0.228 +9\%	& \cellcolor{green!25}\textbf{0.223 +6\%}	& \cellcolor{green!25}0.220 +5\%	& \cellcolor{green!25}\textbf{0.219 +4\%}	& \cellcolor{green!25}0.221 +5\%	& \cellcolor{green!25}0.220 +5\% & \cellcolor{green!25}0.216 +3\%\\
40\%	& \cellcolor{green!25}0.227 +8\%	& \cellcolor{green!25}0.220 +5\%	& \cellcolor{red!25}0.195 -7\%	& \cellcolor{red!25}0.203 -3\%	& \cellcolor{green!25}0.220 +5\%	& \cellcolor{green!25}0.242 +15\% & \cellcolor{green!25}\textbf{0.230 +10\%}\\
50\%	& \cellcolor{green!25}0.230 10\%	& \cellcolor{green!25}0.220 +5\%	& \cellcolor{red!25}0.206 -2\%	& \cellcolor{red!25}0.199 -5\%	& \cellcolor{green!25}\textbf{0.224 +7\%}	& \cellcolor{green!25}0.233 +11\% & \cellcolor{red!25}0.197 -6\%\\
60\%	& \cellcolor{green!25}0.223 +6\%	& \cellcolor{green!25}0.218 +4\%	& \cellcolor{green!25}\textbf{0.221 +5\%}	& \cellcolor{red!25}0.209 -0\%	& \cellcolor{green!25}0.221 +5\%	& \cellcolor{green!25}0.225 +7\% & \cellcolor{green!25}0.211 +0\%\\
70\%	& \cellcolor{green!25}0.223 +6\%	& \cellcolor{green!25}0.215 +2\%	& \cellcolor{green!25}0.213 +1\%	& \cellcolor{red!25}0.208 -1\%	& \cellcolor{green!25}0.211 +0\%	& \cellcolor{green!25}0.249 +19\% & \cellcolor{green!25}0.225 +7\%\\
80\%	& \cellcolor{green!25}0.227 +8\%	& \cellcolor{green!25}0.213 +1\%	& \cellcolor{green!25}0.214 +2\%	& \cellcolor{red!25}0.208 -1\%	& \cellcolor{green!25}0.221 +5\%	& \cellcolor{green!25}0.243 +16\% & \cellcolor{red!25}0.209 -0\%\\
90\%	& \cellcolor{green!25}0.217 +3\%	& \cellcolor{green!25}0.218 +4\%	& \cellcolor{red!25}0.208 -1\%	& \cellcolor{red!25}0.207 -1\%	& \cellcolor{green!25}0.221 +5\%	& \cellcolor{yellow!25}\textbf{0.252 +20\%}   & \cellcolor{red!25}0.200 -5\%\\
100\%	& \cellcolor{green!25}0.230 +10\%	& \cellcolor{green!25}0.222 +6\%	& \cellcolor{red!25}0.209 -0\%	& \cellcolor{green!25}0.214 +2\%	& \cellcolor{green!25}0.215 +2\%	& \cellcolor{green!25}0.235 +12\% & \cellcolor{red!25}0.194 -8\%\\
\bottomrule
\end{tabular}
\end{table}

\subsection{Semi-greedy Enrichment}

The Semi-greedy Enrichment (SemE) is proposed to overcome the randomness of the RanE method and to guarantee an improvement in the classification. The results of SemE are very promising, and by far exceeded our expectations, see Table~\ref{tab_semE} for results obtained with the Random Forest classifier. On average, the classification of the base set was improved by 18\% (bolded cells in Table~\ref{tab_semE}), and in the best cases by impressive 40\% (MIT and Infectious enriched with Twitter observations, yellow cells in Table~\ref{tab_semE}). With the Decision Tree classifier, the improvement was even greater and reached 43\% (see Table~\ref{tab_seme_decision_tree}). The observations from the Twitter data set turned out to be the best enriching subsets among all considered data sets, enhancing the classification of base sets, on average, by 19\%. The reason might be the highest number of observations in the Twitter data set belonging to the minority classes, which gives the SemE algorithm a broader set to pick from. The most susceptible to enrichment base sets were Infectious (on average +28\%) and MIT (on average +21\%). Again, these two data sets are the least numerous, and that can be the reason why they benefited from the new observations more than other data sets.

\begin{table}[H]
\caption{The influence of SemE on the improvement of the classification. The method improved the classification quality, on average, by 18\%. The classification executed with the Random Forest classifier and the F-measure averaged over all classes as the quality measure. Cells marked green indicate an improved classification quality. The best result of each external data set is bolded, and the best overall results are marked yellow.}
\label{tab_semE}
\centering
\setlength\tabcolsep{2pt}
\begin{tabular}{L|cLccccccc}
\toprule
\multirow{2}{1.5cm}{\centering\vfil\bfseries\textbf{External set}}    & \multicolumn{7}{c}{\bfseries Base set}\\ \cmidrule(lr){2-9}  &   \textbf{Digg}	& \textbf{Irvine Messages}	& \textbf{Loans}	& \textbf{Facebook}	& \textbf{MIT}   & \textbf{Slashdot}	& \textbf{Infectious}	& \textbf{Twitter}\\
\midrule
-	& 0.226	& 0.212	& 0.180	& 0.260	& 0.206 & 0.249	& 0.250 & 0.191\\
\textbf{Digg}	& 	& \cellcolor{green!25}0.215 +1\%	& \cellcolor{green!25}0.198 +10\%	&  \cellcolor{green!25}0.267 +3\%	& \cellcolor{green!25}0.230 +12\% & \cellcolor{green!25}0.265 +6\%	&  \cellcolor{green!25}0.334 +34\% &  \cellcolor{green!25}0.201 +5\%\\
\textbf{Irvine Messages}	&  \cellcolor{green!25}0.236 +4\%	& 	&  \cellcolor{green!25}0.208 +16\%	& \cellcolor{green!25}0.263 +1\%	& \cellcolor{green!25}0.228 +11\% & \cellcolor{green!25}0.264 +6\%	& \cellcolor{green!25}0.313 +25\%   &  \cellcolor{green!25}0.198 +4\%\\
\textbf{Loans}	& \cellcolor{green!25}0.231 +2\%	& \cellcolor{green!25}0.218 +3\%	& 	& \cellcolor{green!25}0.261 +0\%	& \cellcolor{green!25}0.238 +16\% & \cellcolor{green!25}0.265 +6\%	& \cellcolor{green!25}0.323 +29\% &  \cellcolor{green!25}0.197 +3\%\\
\textbf{Facebook}	& \cellcolor{green!25}0.228 +1\%	& \cellcolor{green!25}0.217 +2\%	& \cellcolor{green!25}0.198 +10\%	& 	& \cellcolor{green!25}0.245 +19\% & \cellcolor{green!25}\textbf{0.266~+7\%}	& \cellcolor{green!25}0.320 +28\% &  \cellcolor{green!25}\textbf{0.204~+7\%}\\
\textbf{MIT}	& \cellcolor{green!25}0.230 +2\%	& \cellcolor{green!25}0.224 +6\%	& \cellcolor{green!25}0.196 +9\%	& \cellcolor{green!25}0.261 +0\%	&   & \cellcolor{green!25}0.264 +6\%	& \cellcolor{green!25}0.267 +7\% &  \cellcolor{green!25}0.198 +4\%\\
\textbf{Slashdot}	& \cellcolor{green!25}0.239 +6\% & \cellcolor{green!25}\textbf{0.248~+17\%}    & \cellcolor{green!25}\textbf{0.214~+19\%}  & \cellcolor{green!25}0.271 +4\% & \cellcolor{green!25}0.266 +29\% &   & \cellcolor{green!25}0.337 +35\%    & \cellcolor{green!25}0.194 +2\%\\
\textbf{Infectious}	& \cellcolor{green!25}0.233 +3\%	&  \cellcolor{green!25}0.232 +9\%	& \cellcolor{green!25}0.193 +7\%	& \cellcolor{green!25}0.266 +2\%	&  \cellcolor{green!25}0.251 +22\%    & \cellcolor{green!25}0.261 +5\%	&     &  \cellcolor{green!25}0.194 +2\%\\
\textbf{Twitter} & \cellcolor{green!25}\textbf{0.240~+6\%}    & \cellcolor{green!25}0.242 +14\%   & \cellcolor{green!25}0.191 +6\%    & \cellcolor{green!25}\textbf{0.277~+7\%}    & \cellcolor{yellow!25}\textbf{0.288~+40\%} & \cellcolor{green!25}0.265 +6\%  & \cellcolor{yellow!25}\textbf{0.349~+40\%}    & \\
\bottomrule
\end{tabular}
\end{table}

\subsection{Supervised Enrichment}
The Supervised Enrichment (SupE) is more computationally demanding than previous techniques. It may need to run SemE first and then requires assigning the type to all observations from base and external data sets. Table \ref{tab_supE} contains the results of SupE application to eight data sets related to complex networks. The external data set for each base set was determined during the SemE procedure (see Table~\ref{tab_semE}, e.g., the Twitter data set was used for Digg, Facebook, MIT, and Infectious data sets. It can be easily noticed that the additional computational complexity of the SupE method is generously rewarded. The SupE approach raised the classification performance, on average, by 27\%. In the best case, a perfectly balanced Infectious data set, the result was improved by outstanding 66\%. Once more, the smallest data sets, Infectious and MIT, gained most: +66\% and +46\%, respectively.

More detailed results, which include F-measure for each class (see Appendix \ref{appendix:results_by_class}), reveal that the SupE approach almost always improves the results of all classes, minority as well as dominant. There were only a few exceptions where SupE decreased the initial F-measure value of a particular class. It means SupE improves the classification of the minority classes without a negative effect on the other classes, especially the dominant class. This is a highly desired, yet tough to achieve, advantage that has not been demonstrated by any other existing method.

We have also analyzed the influence of the number of external observations added to the base set on the improvement of the classification performance. The amount varies from 0\% (no additional observations, baseline) to 100\% of the difference between $|D_b|$ and $|M_b|$ (to obtain $IR = 1$). One can observe (compare rows in Table \ref{tab_supE}) how the classifier progressively saturates with additional information (borderline observations) and boosts the classification quality up to a certain point. Once at this point, the additional information does not enhance the classifier's ability to differentiate between classes. As a result, further gain in classification performance is not achieved (at least with the considered technique). Interestingly, this point is often met at the very beginning (10\% of $|M_e|-|M_b|$). This phenomenon will be yet investigated. The process of obtaining the external borderline observations is costly; therefore, if one has limited resources, adding only 10\% of additional observations might be an option. In our case, it improved the results of all data sets, on average, by 22\%.
 
%%%%%%%%%%%%%%%%%%%%

\begin{table}[H]
\caption{The influence of SupE on the improvement of classification. The method improved the classification quality on average by 27\%. The classification executed with the Random Forest classifier and the F-measure averaged over all classes as the quality measure. Cells marked green indicate an improved classification quality. The best result of each data set is bolded, and the best overall result is marked yellow.}

\label{tab_supE}
\centering
%% \tablesize{} %% You can specify the fontsize here, e.g., \tablesize{\footnotesize}. If commented out \small will be used.
%\tablesize{\footnotesize}
\setlength\tabcolsep{2pt}
\begin{tabular}{L|cLcccccc}
\toprule
\multirow{2}{1.5cm}{\centering\bfseries\textbf{Amount of $|M_e|-|M_b|$}}    & \multicolumn{8}{c}{\bfseries Base set}\\ \cmidrule(lr){2-9}  &   \textbf{Digg}	& \textbf{Irvine Messages}	& \textbf{Loans}	& \textbf{Facebook}	& \textbf{MIT}  & \textbf{Slashdot}	& \textbf{Infectious}	& \textbf{Twitter}\\
\midrule
0\%	& 0.226	& 0.212	& 0.180	& 0.260	& 0.206 & 0.249	& 0.250 & 0.191\\
10\%	& \cellcolor{green!25}0.265 +17\%	& \cellcolor{green!25}0.250 +18\%	& \cellcolor{green!25}0.218 +21\%	& \cellcolor{green!25}\textbf{0.294 +13\%}	& \cellcolor{green!25}0.278 +35\%   & \cellcolor{green!25}\textbf{0.278 +12\%}	& \cellcolor{green!25}0.370 +48\%	& \cellcolor{green!25}\textbf{0.213 +12\%}\\
20\%	& \cellcolor{green!25}0.266 +18\%	& \cellcolor{green!25}0.251 +18\%	& \cellcolor{green!25}0.219 +22\%	& \cellcolor{green!25}\textbf{0.294 +13\%}	& \cellcolor{green!25}0.286 +39\%   & \cellcolor{green!25}\textbf{0.278 +12\%}	& \cellcolor{green!25}0.367 +47\%	& \cellcolor{green!25}\textbf{0.213 +12\%}\\
30\%	& \cellcolor{green!25}0.267 +18\%	& \cellcolor{green!25}0.250 +18\%	& \cellcolor{green!25}0.221 +23\%	& \cellcolor{green!25}\textbf{0.294 +13\%}	& \cellcolor{green!25}0.284 +38\%   & \cellcolor{green!25}\textbf{0.278 +12\%}	& \cellcolor{green!25}0.373 +49\%	& \cellcolor{green!25}\textbf{0.213 +12\%}\\
40\%	& \cellcolor{green!25}0.267 +18\%	& \cellcolor{green!25}0.252 +19\%	& \cellcolor{green!25}0.219 +22\%	& \cellcolor{green!25}\textbf{0.294 +13\%}	& \cellcolor{green!25}0.284 +38\%   & \cellcolor{green!25}\textbf{0.278 +12\%}	& \cellcolor{green!25}0.370 +48\%	& \cellcolor{green!25}\textbf{0.213 +12\%}\\
50\%	& \cellcolor{green!25}\textbf{0.268 +19\%}	& \cellcolor{green!25}0.252 +19\%	& \cellcolor{green!25}0.223 +24\%	& \cellcolor{green!25}\textbf{0.294 +13\%}	& \cellcolor{green!25}0.288 +40\%   & \cellcolor{green!25}\textbf{0.278 +12\%}	& \cellcolor{green!25}0.382 +53\%	& \cellcolor{green!25}\textbf{0.213 +12\%}\\
60\%	& \cellcolor{green!25}\textbf{0.268 +19\%}	& \cellcolor{green!25}0.254 +20\%	& \cellcolor{green!25}0.223 +24\%	& \cellcolor{green!25}\textbf{0.294 +13\%}	& \cellcolor{green!25}0.289 +40\%   & \cellcolor{green!25}\textbf{0.278 +12\%}	& \cellcolor{green!25}0.375 +50\%	& \cellcolor{green!25}\textbf{0.213 +12\%}\\
70\%	& \cellcolor{green!25}\textbf{0.268 +19\%}	& \cellcolor{green!25}0.256 +21\%	& \cellcolor{green!25}0.224 +25\%	& \cellcolor{green!25}\textbf{0.294 +13\%}	& \cellcolor{green!25}0.291 +41\%   & \cellcolor{green!25}\textbf{0.278 +12\%}	& \cellcolor{green!25}0.387 +55\%	& \cellcolor{green!25}\textbf{0.213 +12\%}\\
80\%	& \cellcolor{green!25}\textbf{0.268 +19\%}	& \cellcolor{green!25}0.258 +22\%	& \cellcolor{green!25}0.225 +25\%	& \cellcolor{green!25}\textbf{0.294 +13\%}	& \cellcolor{green!25}0.293 +42\%   & \cellcolor{green!25}\textbf{0.278 +12\%}	& \cellcolor{green!25}0.396 +58\%	& \cellcolor{green!25}\textbf{0.213 +12\%}\\
90\%	& \cellcolor{green!25}\textbf{0.268 +19\%}	& \cellcolor{green!25}0.257 +21\%	& \cellcolor{green!25}0.228 +27\%	& \cellcolor{green!25}\textbf{0.294 +13\%}	& \cellcolor{green!25}\textbf{0.301 +46\%}  & \cellcolor{green!25}\textbf{0.278 +12\%}	& \cellcolor{green!25}0.405 +62\%	& \cellcolor{green!25}\textbf{0.213 +12\%}\\
100\% $IR=1$	& \cellcolor{green!25}\textbf{0.268 +19\%}	& \cellcolor{green!25}\textbf{0.259 +22\%}	& \cellcolor{green!25}\textbf{0.230 +28\%}	& \cellcolor{green!25}\textbf{0.294 +13\%}	& \cellcolor{green!25}0.300 +46\%   & \cellcolor{green!25}\textbf{0.278 +12\%}	& \cellcolor{yellow!25}\textbf{0.415 +66\%}	& \cellcolor{green!25}\textbf{0.213 +12\%}\\
\bottomrule
\end{tabular}
\end{table}

\subsection{Comparison results}

The proposed approaches were compared with nine state-of-the-art methods for learning from imbalanced data sets: Random Under-sampling, Random Over-sampling, SMOTE~\cite{chawla2002smote}, bSMOTE~\cite{han2005borderline}, SMOTE-ENN~\cite{batista2004study}, SMOTE-TL~\cite{batista2003balancing}, IHT~\cite{smith2014instance}, ADASYN~\cite{he2008adasyn}, MC-RBO~\cite{krawczyk2019radial}. Some are designed to cope with the multi-class classification, and for others, we have used decomposition techniques (OVO and OVA). The detailed setup attributes and hyperparameters adjustments, as well as validation procedure, are provided in Appendix~\ref{appendix:validation}.

Our solutions, namely RanE, SemE, SupE, outperformed the state-of-the-art methods for the multi-class classification task, see Table~\ref{tab_comparison1}. The superiority of the developed algorithms is especially notable for the smallest data sets: Infectious and MIT. For these two sets, all methods from the literature, except bSMOTE, were unable to enhance the classification quality (or achieved negligible improvement), while our solutions provided the best enhancement among all considered data sets.

The RUS was the only method that was unable to improve results for all considered data sets. On average, RUS worsened the results by 23\%. Also, IHT did not perform well, improving only the result for the Twitter data set and worsening the results, on average, by 14\%. The SMOTE-ENN and the SMOTE-TL had difficulty mainly with the smallest data sets, achieving decent results for other data sets. The ADASYN failed to run on data sets with critically low under-represented class; hence, the results are not provided. For other data sets, except Twitter, where it worsened result by 20\%, the ADASYN performed well. The MC-RBO has been applied only to small data sets, as the computations for more massive sets took too long. The MC-RBO's results are indistinctive. The ROS and the SMOTE methods failed only for the smallest data set and the Twitter data set. They gained, on average, 1\% and 2\%, respectively. The best among the compared methods was bSMOTE, which enhanced the results, on average, by 10\%. In the best case, bSMOTE boosted the result by 19\% (the MIT data set). It was the only state-of-the-art method that managed to improve results for all complex network data sets.

Our algorithms, in comparison with the existing solutions, performed significantly better. Even the most simple of them, the RanE method, managed to improve results for all considered data set, even though its principles does not guarantee the improvement. On average, RanE improved the classification quality by 7\%, and in the best case by 17\% (the Infectious data set). The SemE method was even better, achieving, on average, 19\% improvement for all data sets, which is 7\% more than bSMOTE and 10\% more than RanE. Additionally, SemE guarantees that the obtained result will not be lower than the baseline. The SupE approach was by far the best algorithm, increasing the classification capabilities for the considered data sets, on average, by 27\%. This is 16\% more than bSMOTE, 19\% more than RanE, and 8\% more than SemE.

\begin{table}[H]
\caption{The comparison of RanE, SemE, and SupE with the state-of-the-art methods for multi-class classification on eight data sets. The SupE approach is by far the best method. The classification executed with the Random Forest classifier and the F-measure averaged over all classes as the quality measure. Cells marked green indicate an improved classification quality, while cells marked red depict worse results. The best result for each data set is bolded and marked yellow.}
\label{tab_comparison1}
\centering
%% \tablesize{} %% You can specify the fontsize here, e.g., \tablesize{\footnotesize}. If commented out \small will be used.
\begin{tabular}{c|MMMMMMMM|>{\centering\arraybackslash}m{0.05\linewidth}}
\toprule
\multirow{2}{1.5cm}{\centering\vfil\bfseries\textbf{Method}}    & \multicolumn{8}{c}{\bfseries Base set}    & \multirow{2}{0.5cm}{\centering\vfil\bfseries\textbf{Avg gain}}\\ \cmidrule(lr){2-9}  &   \textbf{Digg}	& \textbf{Irvine Messages}	& \textbf{Loans}	& \textbf{Facebook}	& \textbf{MIT}  & \textbf{Slashdot}	& \textbf{Infectious}	& \textbf{Twitter}\\
\midrule
None    & 0.226	& 0.212	& 0.180	& 0.260	& 0.206 & 0.249	& 0.250 & 0.191 & -\\
\textbf{RUS} & \cellcolor{red!25}0.202 -11\% & \cellcolor{red!25}0.186 -12\% & \cellcolor{red!25}0.098 -46\% & \cellcolor{red!25}0.186 -28\% & \cellcolor{red!25}0.132 -36\% & \cellcolor{red!25}0.207 -17\% & \cellcolor{red!25}0.184 -26\% & \cellcolor{red!25}0.174 -9\% & \cellcolor{red!25}-23\%\\
\textbf{ROS} & \cellcolor{green!25}0.227 +0\% & \cellcolor{green!25}0.224 +6\% & \cellcolor{green!25}0.199 +11\% & \cellcolor{green!25}0.261 +0\% & \cellcolor{green!25}0.209 +1\% & \cellcolor{green!25}0.268 +8\% & \cellcolor{red!25}0.236 -6\% & \cellcolor{red!25}0.162 -15\% & \cellcolor{green!25}+1\%\\
\textbf{SMOTE}   & \cellcolor{green!25}0.231 +2\% & \cellcolor{green!25}0.234 +10\% & \cellcolor{green!25}0.201 +12\% & \cellcolor{green!25}0.265 +2\% & \cellcolor{green!25}0.226 +10\% & \cellcolor{green!25}0.271 +9\% & \cellcolor{red!25}0.213 -15\% & \cellcolor{red!25}0.158 -17\% & \cellcolor{green!25}+2\%\\
\textbf{bSMOTE}  & \cellcolor{green!25}0.244 +8\% & \cellcolor{green!25}0.229 +8\% & \cellcolor{green!25}0.193 +7\% & \cellcolor{green!25}0.288 +11\% & \cellcolor{green!25}0.246 +19\% & \cellcolor{yellow!25}\textbf{0.278 +12\%} & \cellcolor{green!25}0.270 +8\% & \cellcolor{green!25}0.204 +7\% & \cellcolor{green!25}+10\%\\
\textbf{SMOTE-ENN}   & \cellcolor{red!25}0.224 -1\% & \cellcolor{green!25}0.222 +5\% & \cellcolor{green!25}0.189 +5\% & \cellcolor{green!25}0.262 +1\% & \cellcolor{red!25}0.167 -19\% & \cellcolor{green!25}0.272 +9\% & \cellcolor{red!25}0.176 -30\% & \cellcolor{green!25}0.193 +1\% & \cellcolor{red!25}-4\%\\
\textbf{SMOTE-TL}    & \cellcolor{green!25}0.238 +5\% & \cellcolor{green!25}0.227 +7\% & \cellcolor{green!25}0.198 +10\% & \cellcolor{green!25}0.286 +10\% & \cellcolor{red!25}0.201 -2\% & \cellcolor{green!25}0.274 +10\% & \cellcolor{red!25}0.247 -1\% & \cellcolor{green!25}0.192 +1\% & \cellcolor{green!25}+5\%\\
\textbf{IHT} & \cellcolor{red!25}0.142 -37\% & \cellcolor{red!25}0.177 -17\% & \cellcolor{red!25}0.155 -14\% & \cellcolor{red!25}0.231 -11\% & \cellcolor{red!25}0.206 +0\% & \cellcolor{red!25}0.191 -23\% & \cellcolor{red!25}0.219 -12\% & \cellcolor{green!25}0.203 +6\% & \cellcolor{red!25}-14\%\\
\textbf{ADASYN} & \cellcolor{green!25}0.245 +8\% & \cellcolor{green!25}0.229 +8\% & \cellcolor{green!25}0.196 +9\% & - & - & \cellcolor{green!25}0.276 +11\% & - & \cellcolor{red!25}0.152 -20\% & -\\
\textbf{MC-RBO} & - & \cellcolor{green!25}0.223 +5\% & - & - & \cellcolor{red!25}0.206 +0\% & - & \cellcolor{green!25}0.261 +5\% & - & -\\
\textbf{RanE}    & \cellcolor{green!25}0.232 +3\% & \cellcolor{green!25}0.228 +8\% & \cellcolor{green!25}0.189 +5\% & \cellcolor{green!25}0.271 +4\% & \cellcolor{green!25}0.233 +13\% & \cellcolor{green!25}0.255 +2\% & \cellcolor{green!25}0.293 +17\% & \cellcolor{green!25}0.195 +2\% & \cellcolor{green!25}+7\%\\
\textbf{SemE}    & \cellcolor{green!25}0.240 +6\% & \cellcolor{green!25}0.248 +17\% & \cellcolor{green!25}0.214 +19\% & \cellcolor{green!25}0.277 +7\% & \cellcolor{green!25}0.288 +40\% & \cellcolor{green!25}0.266 +7\% & \cellcolor{green!25}0.349 +40\% & \cellcolor{green!25}0.204 +7\% & \cellcolor{green!25}+18\%\\
\textbf{SupE}    & \cellcolor{yellow!25}\textbf{0.268 +19\%} & \cellcolor{yellow!25}\textbf{0.259 +22\%} & \cellcolor{yellow!25}\textbf{0.230 +28\%} & \cellcolor{yellow!25}\textbf{0.294 +13\%} & \cellcolor{yellow!25}\textbf{0.301 +46\%} & \cellcolor{yellow!25}\textbf{0.278 +12\%} & \cellcolor{yellow!25}\textbf{0.415 +66\%} & \cellcolor{yellow!25}\textbf{0.213 +12\%} & \cellcolor{yellow!25}\textbf{+27\%}\\
\bottomrule
\end{tabular}
\end{table}

To apply our solutions to the tumor data, we had to select only eight features from the BRE data set that were present in the PRO data set. However, to maintain the completeness of the research, we include the results on the unmodified BRE data set as well, see Table~\ref{tab_cancer}. The classification quality for the BRE data set was the highest when using all 30 features and no balancing algorithm (0.960). Applying any state-of-the-art method on unmodified BRE data set only worsened this result. With only eight features available, the classification quality dropped to 0.911 (-5\%). However, this time, applying our methods (and some methods from the literature) improved the classification performance. In this case, the SupE method turned out to be the best (0.951, +4\%), although the result was still slightly worse than classification on the unmodified BRE data set. Apparently, the features that we have removed contained more learning information than the external observations obtained from the PRO data set. It would be very interesting to apply dimensionality reduction so that all available information could be preserved, but we leave this for future work.

For the PRO data set, all our solutions improved the classification quality. The RanE and SemE methods gained 3\%, while the SupE method gained 6\%. At the same time, only three methods from the literature (SMOTE, bSMOTE, MC-RBO) managed to improve the classification result, and the gain was negligible (less than 1\%).

Overall, our methods surpassed the state-of-the-art methods again, and the SupE method was the best one. Let us note, that this time, the SupE method performed better without applying the SemE step first. Such a situation may happen when the base set already has enough number of minority observations to proceed with SupE. In that case, adding external observations with SemE is unnecessary, as it could move the class decision boundary in the wrong direction.

The obtained results are statistically significant. See Appendix~\ref{appendix:statistical_analysis} for more details, and especially Table~\ref{tab_friedman} for ranks of all compared methods.

\begin{table}[H]
\caption{The comparison of RanE, SemE, and SupE with the state-of-the-art methods for the binary classification on two data sets. To apply RanE, SemE, and SupE, the feature space of the Breast Tumor data set had to be adjusted. The classification executed with the Random Forest classifier and the F-measure averaged over two classes as the quality measure. Cells marked green indicate an improved classification quality, while cells marked red depict worse results. The best result for each data set is bolded and marked yellow.}
\label{tab_cancer}
\centering
\begin{tabular}{c|ccc}
\toprule
\multirow{2}{2cm}{\centering\bfseries\textbf{Balancing method}}    & \multicolumn{3}{c}{\bfseries Base set}\\ \cmidrule(lr){2-4}   & \textbf{BRE (30 features)}  & \textbf{BRE (8 features)}	& \textbf{PRO (8 features)}\\
\midrule
\textbf{None}	& 0.960  & 0.911	& 0.842\\
\textbf{RUS}	& \cellcolor{red!25}0.940 -2\% & \cellcolor{red!25}0.903 -1\%	& \cellcolor{red!25}0.833 -1\%\\
\textbf{ROS}	& \cellcolor{red!25}0.951 -1\% & \cellcolor{green!25}0.923 +1\%	& \cellcolor{red!25}0.834 -1\%\\
\textbf{SMOTE}	& \cellcolor{red!25}0.947 -1\% & \cellcolor{green!25}0.919 +1\%	\cellcolor{green!25}& \cellcolor{green!25}0.849 +1\%\\
\textbf{bSMOTE}	& \cellcolor{red!25}0.948 -1\% & \cellcolor{green!25}0.920 +1\%	& \cellcolor{green!25}0.844 +0\%\\
\textbf{SMOTE-ENN}	& \cellcolor{red!25}0.949 -1\% & \cellcolor{green!25}0.923 +1\%	& \cellcolor{red!25}0.830 -1\%\\
\textbf{SMOTE-TL}	& \cellcolor{red!25}0.948 -1\% & \cellcolor{green!25}0.925 +2\%	& \cellcolor{red!25}0.837 -1\%\\
\textbf{IHT}	& \cellcolor{red!25}0.883 -8\% & \cellcolor{red!25}0.866 -5\%	& \cellcolor{green!25}0.845 +0\%\\
\textbf{ADASYN}	& \cellcolor{red!25}0.944 -2\% & \cellcolor{red!25}0.910 -0\%	& \cellcolor{red!25}0.828 -2\%\\
\textbf{MC-RBO}	& \cellcolor{red!25}0.930 -3\% & \cellcolor{green!25}0.929 +2\%	& \cellcolor{green!25}0.847 +1\%\\
\textbf{RanE}	& - & \cellcolor{green!25}0.923 +1\%	& \cellcolor{green!25}0.868 +3\%\\
\textbf{SemE}	& - & \cellcolor{green!25}0.936 +3\%	& \cellcolor{green!25}0.866 +3\%\\
\textbf{SupE (with SemE)}	& - & \cellcolor{red!25}0.865 -5\%	& \cellcolor{green!25}0.868 +3\%\\
\textbf{SupE (without SemE)}	& - & \cellcolor{yellow!25}\textbf{0.951 +4\%}	& \cellcolor{yellow!25}\textbf{0.894 +6\%}\\
\bottomrule
\end{tabular}
\end{table}

\section{Discussion}

The results indicate that the proposed methods are very successful. There are, however, some points to discuss. First and foremost, the proposed solutions require at least one external data set with the same classes labeled and the same feature space as the base data set. Nevertheless, in the era of vast amounts of data, a necessity of an additional data set is not a significant problem. As of the same feature space requirement, one may perform a dimensionality reduction technique.

On the other hand, the existing methods have some drawbacks too. For example, the under-sampling methods require a certain amount of observations so that the dominant ones can be selected. During this process, some information (observations) is lost. The over-sampling methods, in turn, need a decent number of minority observations in order to produce diverse synthetic observations. Additionally, the observations created in this process are not real but synthetic. Our solution uses the external, but real, observations that could potentially exist in the considered data set.

A fascinating point to address is the idea of enriching the learning set with observations from an external data set, without any context or parametric requirements (besides the same class and feature spaces). With such a flexible principle, one may argue we assume that the characteristic of the observations in base and external data sets are the same. We do not; it is against many conducted research and intuition. We believe that, for instance, groups having the same characteristic may experience different events across various data sets. For example, large groups in co-authorship data set are much more likely to continue existence for many years, rather than to vanish rapidly. At the same time, large discussion groups on social platforms can last just a few hours and dissolve instantly. Nevertheless, our algorithms perform very well, carefree that the additional observations come from a different data set. We have two possible explanations for that. Either networks (or tumors) are not that much different after all, and have a similar pace of evolution (once normalized by the appropriate time window size selection). Alternatively, the processes responsible for the evolution of groups are much more complicated than we thought and not always linked to the data set context. Either way, this phenomenon will be further investigated to understand better which characteristics of the external data set are the critical factors for the successful enrichment of the base data set.

Moreover, the enriching technique, to some extent, can be compared to the transfer learning technique. Although, in our opinion, the influence of the external data context is marginal, as the number of external observations in the learning process is much lower than the number of base observations. The initial experiments revealed that the transfer learning technique performed better than RanE, but worse than the SemE and SupE approaches, on the considered data sets; see Appendix \ref{appendix:transfer_learning}.

Furthermore, the proposed methods can be combined with any other technique for imbalanced data classification. For example, experiments combining SemE and SupE with bSMOTE showed that, in the case of SemE, the final result could be sometimes improved; see Appendix~\ref{appendix:combining_bsmote}.

Lastly, when performing SemE, the order in which the external observations are iterated might influence the final set of selected observations, hence the final classification result. The exact influence of the order will be analyzed in future experiments.

% dodajemy tylko klasy minority, bo chcemy zmniejszac IR

%furthermore, roznice miedzy wynikami poszczegolnych metod pokazuja jak wazne jest dodawanie obiektow brzegowych. 
%(czy inne typy byly sprawdzane?)

%dodac akapit mowiacy o roznicach miedzy metodami, ktorą kiedy uzywac itp. 

%%%%%%%%%%%%%%%%%%%%%%%%%%%%%%%%%%%%%%%%%%
\section{Conclusions and future work}

To overcome the class imbalance problem, we have proposed to enrich the learning set with observations from the external data set. This task can be accomplished with three different approaches: (1) randomly selecting observations (RanE); (2) iteratively choosing observations that improve the classification result (SemE); (3) adding observations that help the classifier to determine the border between classes better (SupE). The developed solutions have been thoroughly validated on eight real-world data sets, and additionally, compared with the state-of-the-art methods.

The conducted experiments revealed that our technique outperformed the state-of-the-art methods, especially for the smallest data sets. The SemE and SupE solutions performed, on average, respectively 7\%, and 16\% better than bSMOTE (which was the best among the methods from the literature). For the two smallest data sets, our approaches were 2\% (RanE), 23\% (SemE), and 39\% (SupE) better than bSMOTE. Apart from the better performance, our technique is free from the drawbacks of the existing methods: (1) no under-sampling means no loss of information, and (2) the lack of synthetic observations avoids bias that can occur when using generated observations. Furthermore, the proposed approaches can be used together with other data balancing methods, as well as any classification algorithm aimed at the class imbalance problem.

Among the presented approaches, SupE is by far the best one, achieving, on average, 27\% improvement in classification quality. In the best case, the SupE method was able to enhance the final result by outstanding 66\%. Even more remarkable is a unique ability of SupE to improve the classification performance on the minority classes without decreasing the performance of the dominant classes. We believe this is thanks to the well-thought-out process of external observations selection, in which only the particular type of information is acquired, i.e., the borderline observations, which help classifier to differentiate classes.

Let us also emphasize that the developed enrichment technique can be successfully applied to the multi-class classification task, for which the number of existing methods dealing with CIP is limited.

To further contribute to the field, we will continue to analyze which data sets' characteristics should be similar to achieve the highest possible gain in classification quality. We will also perform the dimensionality reduction experiments (such as feature embedding or principal component analysis), to prove that different feature space is not a limitation when choosing the external set. Finally, we will determine the influence of the order in which the external observations are added in the SemE approach on the final result.

\subsection*{Funding}
This work was partially supported by the National Science Centre, Poland, project no. 2016/21/B/ST6/01463; and the statutory funds of the Department of Computational Intelligence, Wroclaw University of Science and Technology.

%%%%%%%%%%%%%%%%%%%%%%%%%%%%%%%%%%%%%%%%%%
\subsection*{Acknowledgments}
Authors thank Michał Koziarski for his consultations during the experiments and manuscript creation, as well as for sharing the code of the MC-RBO method. Authors also thank the Data Science Group at Wrocław University of Science and Technology for the discussion that improved this work.

\bibliographystyle{unsrt}  
\bibliography{bibliography}

\newpage

\appendix

\section{Experimental setup}
\label{appendix:experiment_setup}

\subsection{Data sets}
\label{appendix:data_sets}

Eight data sets on real-world complex networks were used for the research: Digg, Irvine Messages, Loans, Facebook, MIT, Slashdot, Infectious, and Twitter. Each data set was described using 88 features, where six classes (event types) were distinguished, in order to predict group changes. The process of transforming the temporal network into the groups described by features was carried out using the Group Evolution Prediction (GEP) method \cite{saganowski2019gep}. All data sets are publicly available \cite{saganowski2018data}. Each data set has a different number of observations and a different class imbalance ratio. The imbalance ratio of each data set is presented in Figure \ref{datasets_IR} and in Appendix \ref{appendix:results_by_class}.

Additionally, two tumor data sets have been used, one related to breast tumor (BRE) \cite{dataset_BRE}, and the other one related to prostate tumor (PRO) \cite{dataset_PRO}. The BRE data set included 30 features, while the PRO data set only 8. Therefore, we have selected from the BRE data only features that were present in the PRO data.

\begin{figure}[H]
\centering
\includegraphics[width=0.7\linewidth]{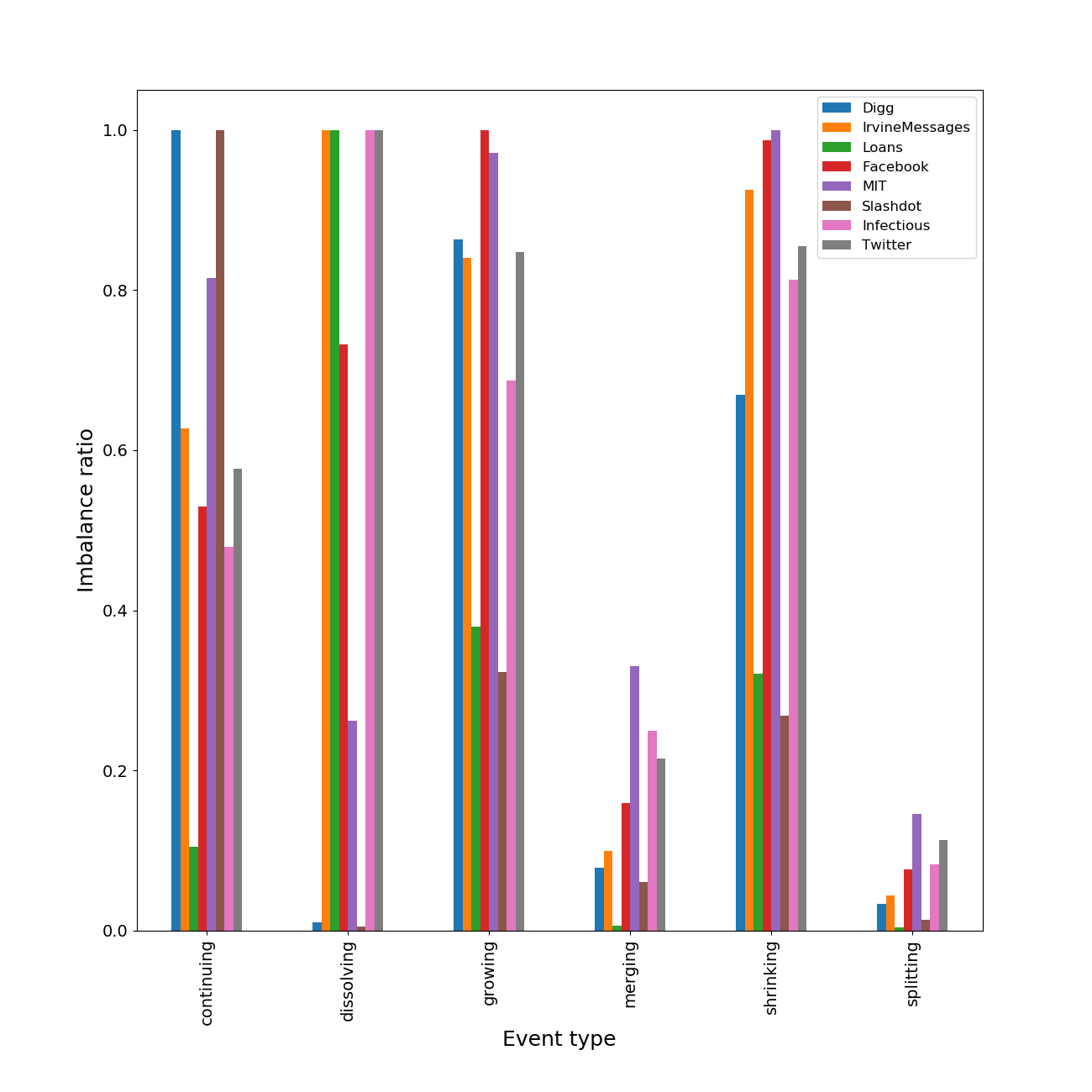}
\caption{Distribution of imbalance ratio in the surveyed datasets.}
\label{datasets_IR}
\end{figure}

\subsection{Classifiers used}
\label{appendix:classifiers_used}

The study selected seven classifiers: AdaBoost (AB), Decision Tree (DT), k-nearest neighbors classifier (kNN), Random Forest (RF), Support Vector Machine (SVM), Naive Bayes classifier (NB), and Neural Network (NN), as these are the most often used in the CIP \cite{haixiang2017learning}. Each of them has been implemented using the scikit-learn package. Parameters of each classifier were constant between data sets. Each of the parameters was selected sequentially based on 10-fold cross-validation, by averaging the results obtained on all examined data sets repeated ten times. Classifier parameters were as follows:
\begin{itemize}
\item   k-nearest neighbors classifier
\begin{itemize}
\item   Number of neighbors: 3
\item   Distance metric: minkowski
\item   Weight function used in prediction: uniform
\item   Other parameters such as the default values for KNeighborsClassifier in scikit-learn package
\end{itemize}
\item   Support Vector Machine
\begin{itemize}
\item   Penalty parameter C of the error term: 0.001
\item   Kernel type: linear
\item   Hard limit on iterations within solver: 1000
\item   Tolerance for stopping criterion: 0.001
\item   Other parameters such as the default values for SVC in scikit-learn package
\end{itemize}
\item   Decision Tree
\begin{itemize}
\item   Criterion: gini
\item   The minimum number of samples required to split an internal node: 2
\item   The minimum number of samples required to be at a leaf node: 1
\item   The maximum depth of the tree: 30
\item   Other parameters such as the default values for DecisionTreeClassifier in scikit-learn package
\end{itemize}
\item   Random Forest
\begin{itemize}
\item   Criterion: gini
\item   The number of trees in the forest: 10
\item   The minimum number of samples required to split an internal node: 2
\item   The minimum number of samples required to be at a leaf node: 1
\item   The maximum depth of the tree: 30
\item   Other parameters such as the default values for RandomForestClassifier in scikit-learn package
\end{itemize}
\item   Neural Network
\begin{itemize}
\item   Number of hidden layers: 1
\item   Number of neurons in hidden layers: 150
\item   Activation function: ReLU
\item   Optimizer: Adam
\item   Initial learning rate: 0.001
\item   L2 penalty: 0.0001
\item   Maximum number of iterations: 1000
\item   Other parameters such as the default values for MLPClassifier in scikit-learn package
\end{itemize}
\item   AdaBoost
\begin{itemize}
\item   Base estimator: Decision Tree
\item   Number of estimators: 10
\item   Learning rate: 1.0
\item   Other parameters such as the default values for AdaBoostClassifier in scikit-learn package
\end{itemize}
\item   AdaBoost
\begin{itemize}
\item   Parameters such as the default values for GaussianNB in scikit-learn package
\end{itemize}
\end{itemize}

\subsection{Classification quality measure}
\label{appendix:classification_measure}

The most commonly used measures to determine the evaluation achievements, adapted to the problem of unbalanced data, include \cite{haixiang2017learning}: Receiver Operating Characteristics, G-Mean, and F-measure. This is due to the reduced susceptibility to imbalanced data distribution, caused by the inclusion of class distribution in the calculation of the measure. To determine the effectiveness of the classification in our study, the F-measure averaged over all classes is used. The plain average over all classes is equally affected by all classes; thus, very well reflects the overall quality of the predictions of the considered data sets. Additional arguments for using the F-measure averaged over all classes are stated in \cite{saganowski2019gep}.

\subsection{Validation}
\label{appendix:validation}

We have applied the stratified 10-fold cross-validation method for the validation process. The small exception to this was the Infectious data set, for which a stratified 4-fold cross-validation has been applied, as there were only four observations in the minority class.

To eliminate the randomness of some methods, experiments were repeated many times, and the results were averaged. The detailed validation process for each considered technique was as follows:
\begin{itemize}
\item    Base score, RUS, ROS - experiments were repeated 50 times, and the final result was the average of all attempts.
\item    SMOTE, bSMOTE, SMOTE-ENN, SMOTE-TL, ADASYN - at first the number of nearest neighbors parameter was adjusted. For each value from the set \{1, 2, ..., 10\} 10 classification attempts were performed, and results were averaged. The value that resulted in the best classification quality was selected for the experiment. The experiment was repeated 50 times, and the final result was the average of all attempts. For each data set, this procedure was conducted separately.
\item MC-RBO - similarly to SMOTE, but the best value for the number of nearest neighbors parameter was selected from the set \{1, 2, ..., 10, 15, 20, ..., 50\}. The number of attempts was the same.
\item    IHT - averaging 50 attempts to balance the data set with the Random Forest as the estimator and 10-fold cross-validation
\end{itemize}

\begin{figure}[H]
\centering
\includegraphics[width=0.7\linewidth]{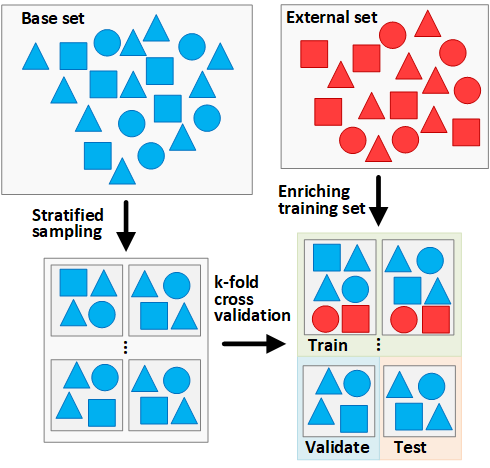}
\caption{The validation procedure. The external observations are added only to the training set.}
\label{fig_enriching_procedure}
\end{figure}

Let us also emphasize that in the case of RanE, SemE, and SupE, the external observations are added only to the training set, see Fig~\ref{fig_enriching_procedure}. The test set and the validation set (used only in SemE) are kept untouched, i.e., contain only observations from the base set.

\subsection{Reproducibility}
\label{appendix:reproducibility}

The implementations of methods used in the experiments are publicly available at:
\begin{itemize}
\item   RanE, SemE, and SupE methods: \url{https://github.com/HJegierski/DataEnrichment}
\item   MC-RBO method: \url{https://github.com/michalkoziarski/MultiClassRBO/blob/master/algorithms.py}
\item   Other state-of-the-art data balancing methods: \url{https://imbalanced-learn.readthedocs.io/en/stable/index.html}
\item   Classifiers: \url{https://scikit-learn.org/stable/}
\end{itemize}

\section{RanE additional results}
\label{appendix:rane_full_results}

The study of the influence of RanE on the correctness of prediction was carried out for all considered data sets; however, the MIT set was the most susceptible to the influence of the developed method. Table \ref{tab_rane_classifiers} shows the influence of the enrichment amount and the classifier selected, on the classification quality enhancement. The first row is the reference point and determines the correctness of the classifier prediction without enrichment. The green color marks cases where the enrichment positively affected the improvement of prediction, while red marks preservation or deterioration of prediction. The best value for each classifier is highlighted in bold font, while the yellow color marks the best performance among all classifiers.

The top three classifiers were the Decision Tree (0.252), AdaBoost (0.243), and Random Forests (0.233). The enrichment allowed to increase the correctness of the classification of each classifier, where the highest profit was in the case of the Neural Networks, reaching 24\%. Let us emphasize that for each classifier, there was an enrichment set that improved the prediction quality.

\begin{table}[H]
\caption{The RanE method performed for the MIT data set with the Infectious as the external data set. A different enrichment amount and various classifiers are analyzed. The F-measure averaged over all classes is used as the quality measure. Cells marked green indicate an improved classification quality, while cells marked red depict worse results. The best result of each classifier is bolded, and the best overall result is marked yellow.}

\label{tab_rane_classifiers}
\centering
%% \tablesize{} %% You can specify the fontsize here, e.g., \tablesize{\footnotesize}. If commented out \small will be used.
\begin{tabular}{c|MMMMMMM}
\toprule
\multirow{2}{2cm}{\centering\bfseries\textbf{Enrichment amount}}    & \multicolumn{7}{c}{\bfseries Classifier}\\ \cmidrule(lr){2-8}  &   \textbf{kNN}	& \textbf{SVM}	& \textbf{DT}	& \textbf{RF}	& \textbf{NN}	& \textbf{AB}	& \textbf{NB}\\
\midrule
0\%	& 0.182	& 0.180	& 0.210	& 0.206	& 0.193	& 0.225	& 0.127\\
10\%	& \cellcolor{green!25}\textbf{0.187 +3\%}	& \cellcolor{red!25}0.165 -8\%	& \cellcolor{green!25}0.233 +11\%	& \cellcolor{red!25}0.199 -3\%	& \cellcolor{green!25}0.217 +12\%	& \cellcolor{red!25}0.207 -8\%	& \cellcolor{green!25}0.150 +18\%\\
20\%	& \cellcolor{red!25}0.177 -3\%	& \cellcolor{red!25}0.180 +0\%	& \cellcolor{green!25}0.215 +2\%	& \cellcolor{red!25}0.205 -0\% 	& \cellcolor{green!25}0.226 +17\%	& \cellcolor{green!25}\textbf{0.243 +8\%}	& \cellcolor{green!25}0.151 +19\%\\
30\%	& \cellcolor{red!25}0.168 -8\%	& \cellcolor{red!25}0.169 -6\%  & \cellcolor{green!25}0.220 +5\%	& \cellcolor{green!25}0.210 +2\%	& \cellcolor{green!25}\textbf{0.239 +24\%}	& \cellcolor{red!25}0.221 -2\%	& \cellcolor{green!25}0.154 +21\%\\
40\%	&\cellcolor{red!25} 0.169 -7\%	& \cellcolor{red!25}0.174 -3\%	& \cellcolor{green!25}0.242 +15\%	& \cellcolor{green!25}\textbf{0.233 +13\%}	& \cellcolor{green!25}0.222 +15\%	& \cellcolor{red!25}0.222 -1\%	& \cellcolor{green!25}\textbf{0.171 +35\%}\\
50\%	& \cellcolor{red!25}0.168 -8\%	& \cellcolor{red!25}0.167 -7\%	& \cellcolor{green!25}0.233 +11\%	&\cellcolor{green!25} 0.209 +1\%	& \cellcolor{green!25}0.221 +15\%	& \cellcolor{green!25}0.226 +0\%	& \cellcolor{green!25}\textbf{0.171 +35\%}\\
60\%	& \cellcolor{red!25}0.168 -8\%   &\cellcolor{red!25} 0.175 -3\%	& \cellcolor{green!25}0.225 +7\%	& \cellcolor{green!25}0.208 +1\%	& \cellcolor{green!25}0.221 +15\%	& \cellcolor{red!25}0.216 -4\%	& \cellcolor{green!25}\textbf{0.171 +35\%}\\
70\%	& \cellcolor{red!25}0.168 -8\%	& \cellcolor{red!25}0.168 -7\%	& \cellcolor{green!25}0.249 +19\%	& \cellcolor{red!25}0.199 -3\%	& \cellcolor{green!25}0.217 +12\%	& \cellcolor{green!25}0.226 +0\%	& \cellcolor{green!25}\textbf{0.171 +35\%}\\
80\%	& \cellcolor{red!25}0.168 -8\%	& \cellcolor{red!25}0.168 -7\%	& \cellcolor{green!25}0.243 +16\%	& \cellcolor{red!25}0.203 -1\%	& \cellcolor{green!25}0.232 +20\%	& \cellcolor{red!25}0.195 -13\%	& \cellcolor{green!25}\textbf{0.171 +35\%}\\
90\%	& \cellcolor{red!25}0.168 -8\%	& \cellcolor{red!25}0.164 -9\%	& \cellcolor{yellow!25}\textbf{0.252 +20\%}	& \cellcolor{red!25}0.197 -4\%	& \cellcolor{green!25}0.222 +15\%	& \cellcolor{red!25}0.215 -4\%	& \cellcolor{green!25}\textbf{0.171 +35\%}\\
100\%	& \cellcolor{red!25}0.168 -8\%	& \cellcolor{green!25}\textbf{0.181 +1\%}	& \cellcolor{green!25}0.235 +12\%	& \cellcolor{green!25}0.219 +6\%	& \cellcolor{green!25}0.221 +15\%	& \cellcolor{green!25}0.232 +3\%	& \cellcolor{green!25}\textbf{0.171 +35\%}\\
\bottomrule
\end{tabular}
\end{table}

\section{SemE additional results}
\label{appendix:seme_full_results}

\begin{table}[H]
\caption{The influence of SemE on the improvement of classification. The method improved the classification quality by up to 18\%. The classification executed with the Decision Tree classifier and the F-measure averaged over all classes as the quality measure. Cells marked green indicate an improved classification quality. The best result of each external data set is bolded, and the best overall results are marked yellow.}
\label{tab_seme_decision_tree}
\centering
%% \tablesize{} %% You can specify the fontsize here, e.g., \tablesize{\footnotesize}. If commented out \small will be used.
\begin{tabular}{L|NNNNNNN}
\toprule
\multirow{2}{1.5cm}{\centering\vfil\bfseries\textbf{Enrichment set}}    & \multicolumn{7}{c}{\bfseries Base set}\\ \cmidrule(lr){2-8}  &   \textbf{Digg}	& \textbf{Irvine Messages}	& \textbf{Loans}	& \textbf{Facebook}	& \textbf{MIT}	& \textbf{Infectious}	& \textbf{Twitter}\\
\midrule
None	& 0.219	& 0.220	& 0.188	& 0.255	& 0.210	& 0.278 & 0.183\\
Digg	& 	&  \cellcolor{green!25}0.226 +3\%	&  \cellcolor{green!25}0.197 +5\%	&  \cellcolor{green!25}0.256 +0\%	&   \cellcolor{green!25}0.265 +26\%	&  \cellcolor{green!25}0.326  +17\% 	&  \cellcolor{green!25}0.194 +6\%\\
Irvine Messages	&  \cellcolor{green!25}0.231 +5\%	& 	&  \cellcolor{green!25}0.204 +9\%	&  \cellcolor{green!25}0.265 +4\%	&  \cellcolor{green!25}0.245 +17\%	&  \cellcolor{green!25}0.332 +19\% 	&  \cellcolor{green!25}0.195 +7\%\\
Loans	&  \cellcolor{green!25}0.221 +1\%	&  \cellcolor{green!25}0.230 +5\%	& 	&  \cellcolor{green!25}0.262 +3\%	&  \cellcolor{green!25}0.234 +11\%	&  \cellcolor{green!25}0.298 +7\% 	&  \cellcolor{green!25}0.192 +5\%\\
Facebook	&  \cellcolor{green!25}0.231 +5\%	&   \cellcolor{green!25}0.236 +7\%	&   \cellcolor{green!25}0.213 +13\%	& 	&  \cellcolor{green!25}0.240 +14\%	&  \cellcolor{green!25}0.285 +3\% 	&  \cellcolor{yellow!25}\textbf{0.197 +8\%}\\
MIT	&  \cellcolor{green!25}0.223 +2\%	&  \cellcolor{green!25}0.225 +2\%	&  \cellcolor{green!25}0.197 +5\%	&  \cellcolor{green!25}0.258 +1\%	& 	&   \cellcolor{green!25}0.340 +22\% 	&  \cellcolor{green!25}0.190 +4\%\\
Infectious	&   \cellcolor{green!25}0.232 +6\%	&  \cellcolor{green!25}0.230 +5\%	&  \cellcolor{green!25}0.192 +2\%	&   \cellcolor{green!25}0.266 +4\%	&  \cellcolor{green!25}0.253 +20\%	&  	&  \cellcolor{green!25}0.191 +4\%\\
Twitter & \cellcolor{yellow!25}\textbf{0.238 +9\%}    & \cellcolor{yellow!25}\textbf{0.251 +14\%}   & \cellcolor{yellow!25}\textbf{0.214 +14\%}    & \cellcolor{yellow!25}\textbf{0.271 +6\%}    & \cellcolor{yellow!25}\textbf{0.300 +43\%}  & \cellcolor{yellow!25}\textbf{0.346 +24\%}    & \\
\bottomrule
\end{tabular}
\end{table}

\section{Statistical analysis}
\label{appendix:statistical_analysis}

The comparison of the average F-measures obtained by $N$ methods for $M$ data sets was carried out according to the Friedman rank test \cite{friedman1937use}, which is used to detect differences between more than two variables in many test samples. The result of such a test only determines the occurrence of differences between the averages compared; therefore, an ad-hoc comparison should be carried out, which will indicate pairs of averages that differ in a statistically significant way. For this purpose, Shaffer's post-hoc multiple comparisons test \cite{shaffer1986modified} was used. Non-parametric statistical analysis was calculated using the KEEL tool \cite{alcala2009keel}.

The Friedman procedure was carried out on the results presented in Tables \ref{tab_comparison1} and \ref{tab_cancer} (BRE with eight features and PRO). This includes all ten data sets considered in our work. The average ranks obtained with the Friedman test are presented in Table \ref{tab_friedman}. The test determined p-value=$4.92*10^{-10}$, which is much lower than 0.05. Thus, the obtained value confirms the statistical significance of the obtained results. As expected, SupE was ranked first, followed by the SemE, bSMOTE, and RanE methods. The results of the Shaffer's multiple post-hoc comparisons for $\alpha=0.05$ are presented in Table \ref{tab_shaffer}; only the statistically significant comparisons are included. The comparison showed that the differences between the developed methods and no balancing of the set were statistically significant, in contrast to the other methods, the differences of which compared to the lack of balancing were not statistically significant. This confirms the positive influence of the developed techniques on the quality of classification.

\begin{table}[H]
\caption{The average ranks of data set balancing algorithms obtained using the Friedman test.}
\label{tab_friedman}
\centering
%% \tablesize{} %% You can specify the fontsize here, e.g., \tablesize{\footnotesize}. If commented out \small will be used.
\begin{tabular}{cc}
\toprule
\textbf{Data set balancing algorithm}	& \textbf{Rank}\\
\midrule
SupE & 1.15\\
SemE & 3.05\\
bSMOTE & 3.9\\
RanE & 5.05\\
SMOTE-TL & 5.5\\
SMOTE & 6.1\\
ROS & 6.3\\
SMOTE-ENN & 7.85\\
None & 8.05\\
IHT & 8.85\\
RUS & 10.2\\
\bottomrule
\end{tabular}
\end{table}

\begin{table}[H]
\caption{The Shaffer post-hoc multiple comparisons of different balancing datasets algorithms for $\alpha=0.05$. Only the hypotheses with p-value above $0.05$ are presented.}
\label{tab_shaffer}
\centering
%% \tablesize{} %% You can specify the fontsize here, e.g., \tablesize{\footnotesize}. If commented out \small will be used.
\begin{tabular}{cc}
\toprule
\textbf{Dataset balancing algorithm}	& \textbf{p value}\\
\midrule
SupE vs. RUS&0\\
SupE vs. IHT&0\\
SemE vs. RUS&0.000001\\
SupE vs. None&0.000003\\
SupE vs. SMOTE-ENN&0.000006\\
bSMOTE vs. RUS&0.000022\\
SemE vs. IHT&0.000092\\
RanE vs. RUS&0.000516\\
SupE vs. ROS&0.000516\\
SemE vs. None&0.000749\\
bSMOTE vs. IHT&0.000846\\
SupE vs. SMOTE&0.000846\\
SemE vs. SMOTE-ENN&0.001211\\
RUS vs. SMOTE-TL&0.001531\\
SupE vs. SMOTE-TL&0.00336\\
bSMOTE vs. None&0.005143\\
RUS vs. SMOTE&0.005706\\
bSMOTE vs. SMOTE-ENN&0.007743\\
RUS vs. ROS&0.008554\\
SupE vs. RanE&0.008554\\
RanE vs. IHT&0.010408\\
SMOTE-TL vs. IHT&0.02391\\
SemE vs. ROS&0.028441\\
SemE vs. SMOTE&0.039753\\
RanE vs. None&0.043114\\
\bottomrule
\end{tabular}
\end{table}

\section{Combining enriching with bSMOTE}
\label{appendix:combining_bsmote}

To evaluate if our solutions can be successfully combined with other balancing methods, we have mixed SemE and SupE with bSMOTE. The bSMOTE method has been chosen as it performed best among the examined state-of-the-art methods. Table~\ref{tab_seme_smote} contains the results of combining SemE with bSMOTE. Please note that percentages and cells' colors in Table~\ref{tab_seme_smote} are calculated with respect to Table~\ref{tab_semE} (the sole SemE), not the baseline (the classification result without the balancing algorithm). Therefore, it is visible that, in most cases, the bSMOTE further improved the result of SemE.

On the contrary, mixing SupE with bSMOTE has significantly worsened the results with respect to the sole SupE results; see Table~\ref{tab_enr_bsmote}. Such results suggest that enriching the base set with the SemE approach leaves some space for establishing a better border between classes. This is possible because a single iteration (semi-greedy) over the external observations does not exhaust the full potential of external information (observations). Meanwhile, the sole SupE (with extra SemE step if necessary) helps to establish a well-positioned border. Therefore, adding bSMOTE on top of this, only results in over-fitting, as there are too many examples near the border.

\begin{table}[H]
\caption{Mixing SemE with bSMOTE. Note that percentages and cells' colors are calculated with respect to Table~\ref{tab_semE} (the sole SemE), not the baseline (the first row). The classification executed with the Random Forest classifier and the F-measure averaged over all classes as the quality measure.}
\label{tab_seme_smote}
\centering
%% \tablesize{} %% You can specify the fontsize here, e.g., \tablesize{\footnotesize}. If commented out \small will be used.
\begin{tabular}{L|MMMMMMMM}
\toprule
\multirow{2}{1.5cm}{\centering\vfil\bfseries\textbf{External set}}    & \multicolumn{7}{c}{\bfseries Base set}\\ \cmidrule(lr){2-9}  &   \textbf{Digg}	& \textbf{Irvine Messages}	& \textbf{Loans}	& \textbf{Facebook}	& \textbf{MIT}   & \textbf{Slashdot}	& \textbf{Infectious}	& \textbf{Twitter}\\
\midrule
\textbf{None}	& 0.226	& 0.212	& 0.180	& 0.260	& 0.206 & 0.249	& 0.250 & 0.191\\
\textbf{Digg}	& 	& \cellcolor{green!25}0.239 +11\%	& \cellcolor{red!25}0.198 +0\%	& \cellcolor{green!25}0.292 +9\%	& \cellcolor{green!25}0.244 +6\%	& \cellcolor{red!25}0.265 +0\%	& \cellcolor{green!25}0.335 +0\%	& \cellcolor{green!25}0.204 +1\%\\
\textbf{Irvine Messages}	& \cellcolor{green!25}0.246 +4\%	& 	& \cellcolor{red!25}0.199 -4\%	& \cellcolor{green!25}0.293 +11\%	& \cellcolor{green!25}0.275 +21\%	& \cellcolor{red!25}0.264 +0\%	& \cellcolor{green!25}0.361 +15\%	& \cellcolor{red!25}0.196 -1\%\\
\textbf{Loans}	& \cellcolor{green!25}0.253 +10\%	& \cellcolor{green!25}0.231 +6\%	& 	& \cellcolor{green!25}0.293 +12\%	& \cellcolor{green!25}0.284 +19\%	& \cellcolor{green!25}0.268 +1\%	& \cellcolor{green!25}0.325 +1\%	& \cellcolor{green!25}0.199 +1\%\\
\textbf{Facebook}	& \cellcolor{green!25}0.247 +8\%	& \cellcolor{green!25}0.247 +14\%	& \cellcolor{green!25}0.199 +1\%	& 	& \cellcolor{green!25}0.288 +18\%	& \cellcolor{green!25}0.271 +2\%	& \cellcolor{green!25}0.351 +10\%	& \cellcolor{red!25}0.200 -2\%\\
\textbf{MIT}	& \cellcolor{green!25}0.249 +8\%	& \cellcolor{green!25}0.232 +4\%	& \cellcolor{green!25}0.207 +6\%	& \cellcolor{green!25}0.292 +12\%	& 	& \cellcolor{green!25}0.271 +3\%	& \cellcolor{green!25}0.340 +27\%	& \cellcolor{green!25}0.205 +4\%\\
\textbf{Slashdot}	& \cellcolor{green!25}0.244 +2\%	& \cellcolor{green!25}0.250 +1\%	& \cellcolor{green!25}0.217 +1\%	& \cellcolor{green!25}0.272 +0\%	& \cellcolor{red!25}0.241 -9\%	& 	& \cellcolor{red!25}0.322 -4\%	& \cellcolor{green!25}0.199 +3\%\\
\textbf{Infectious}	& \cellcolor{green!25}0.248 +6\%	& \cellcolor{green!25}0.244 +5\%	& \cellcolor{green!25}0.203 +5\%	& \cellcolor{green!25}0.292 +10\%	& \cellcolor{green!25}0.252 +0\%	& \cellcolor{green!25}0.264 +1\%	& 	& \cellcolor{green!25}0.200 +3\%\\
\textbf{Twitter}	& \cellcolor{green!25}0.244 +2\%	& \cellcolor{red!25}0.232 -4\%	& \cellcolor{green!25}0.198 +4\%	& \cellcolor{green!25}0.293 +6\%	& \cellcolor{red!25}0.270 -6\%	& \cellcolor{red!25}0.261 -2\%	& \cellcolor{green!25}0.360 +3\%	& \\
\bottomrule
\end{tabular}
\end{table}

\begin{table}[H]
\caption{Mixing SemE and SupE with bSMOTE. The classification executed with the Random Forest classifier and the F-measure averaged over all classes as the quality measure. Cells marked green indicate an improved classification quality, while cells marked red depict worse results. The best result for each data set is bolded and marked yellow.}
\label{tab_enr_bsmote}
\centering
%% \tablesize{} %% You can specify the fontsize here, e.g., \tablesize{\footnotesize}. If commented out \small will be used.
\begin{tabular}{L|MMMMMMMM}
\toprule
\multirow{2}{2cm}{\centering\vfil\centering\bfseries\textbf{Balancing method}}    & \multicolumn{7}{c}{\bfseries Base set}\\ \cmidrule(lr){2-9}  &   \textbf{Digg}	& \textbf{Irvine Messages}	& \textbf{Loans}	& \textbf{Facebook}	& \textbf{MIT}   & \textbf{Slashdot}	& \textbf{Infectious}	& \textbf{Twitter}\\
\midrule
\textbf{None}	& 0.226	& 0.212	& 0.180	& 0.260	& 0.206 & 0.249	& 0.250 & 0.191\\
\textbf{SemE}    & \cellcolor{green!25}0.240 +6\% & \cellcolor{green!25}0.248 +17\% & \cellcolor{green!25}0.214 +19\% & \cellcolor{green!25}0.277 +7\% & \cellcolor{green!25}0.288 +40\% & \cellcolor{green!25}0.266 +7\% & \cellcolor{green!25}0.349 +40\% & \cellcolor{green!25}0.204 +7\%\\
\textbf{SemE + bSMOTE}	& \cellcolor{green!25}0.253 +12\%	& \cellcolor{green!25}0.250 +18\%	& \cellcolor{green!25}0.217 +21\%	& \cellcolor{green!25}0.293 +13\%	& \cellcolor{green!25}0.288 +40\%	& \cellcolor{green!25}0.271 +9\%	& \cellcolor{green!25}0.361 +44\%	& \cellcolor{green!25}0.205 +7\%\\
\textbf{SupE}   & \cellcolor{yellow!25}\textbf{0.268 +19\%} & \cellcolor{yellow!25}\textbf{0.259 +22\%} & \cellcolor{yellow!25}\textbf{0.230 +28\%} & \cellcolor{yellow!25}\textbf{0.294 +13\%} & \cellcolor{yellow!25}\textbf{0.301 +46\%} & \cellcolor{yellow!25}\textbf{0.278 +12\%} & \cellcolor{yellow!25}\textbf{0.415 +66\%} & \cellcolor{yellow!25}\textbf{0.213 +12\%}\\
\textbf{SupE + bSMOTE}	& \cellcolor{green!25}0.247 +9\%	& \cellcolor{green!25}0.236 +11\%	& \cellcolor{green!25}0.190 +6\%	& \cellcolor{yellow!25}\textbf{0.294 +13\%}	& \cellcolor{green!25}0.246 +19\%	& \cellcolor{green!25}0.261 +5\%	& \cellcolor{green!25}0.321 +28\%	& \cellcolor{green!25}0.198 +4\%\\
\bottomrule
\end{tabular}
\end{table}

\section{Classification results by class}
\label{appendix:results_by_class}

% Table generated by Excel2LaTeX from sheet 'by class to art'
\begin{table}[H]
  \centering
  \caption{The comparison of RanE, SemE, and SupE with the state-of-the-art methods for multi-class
classification on Digg dataset. The classification executed with the Random Forest classifier and the F-measure averaged over all classes as the quality measure.}
  \setlength\tabcolsep{4pt}
    \begin{tabular}{l|rrrrrrrrrrrr|MO}
    \multicolumn{15}{c}{\textbf{Digg}} \\
    \midrule
    \multicolumn{1}{r}{} & \multicolumn{12}{c|}{Class label}                                                             & \multicolumn{1}{r}{\multirow{4}{0.07\linewidth}{Avg [F-me asure]}} & \multicolumn{1}{c}{\multirow{4}{0.04\linewidth}{Gain [\%]}} \\
    \multicolumn{1}{r}{} & \multicolumn{2}{c}{Continuing} & \multicolumn{2}{c}{Dissolving} & \multicolumn{2}{c}{Growing} & \multicolumn{2}{c}{Merging} & \multicolumn{2}{c}{Shrinking} & \multicolumn{2}{c|}{Splitting} &       &  \\
\cmidrule{1-13}    Count & \multicolumn{2}{c}{6371} & \multicolumn{2}{c}{64} & \multicolumn{2}{c}{5503} & \multicolumn{2}{c}{504} & \multicolumn{2}{c}{4261} & \multicolumn{2}{c|}{213} &       &  \\
    IR    & \multicolumn{2}{c}{1.0} & \multicolumn{2}{c}{99.5} & \multicolumn{2}{c}{1.2} & \multicolumn{2}{c}{12.6} & \multicolumn{2}{c}{1.5} & \multicolumn{2}{c|}{29.9} &       &  \\
    \midrule
    None  & 0.490 & \multicolumn{1}{c}{-} & 0.000 & \multicolumn{1}{c}{-} & 0.366 & \multicolumn{1}{c}{-} & 0.020 & \multicolumn{1}{c}{-} & 0.482 & \multicolumn{1}{c}{-} & 0.000 & \multicolumn{1}{c|}{-} & 0.226 & \multicolumn{1}{c}{-} \\
    RUS   & \cellcolor{red!25}0.384 & \cellcolor{red!25}-22\% & \cellcolor{green!25}0.022 & \cellcolor{green!25}NaN & \cellcolor{red!25}0.263 & \cellcolor{red!25}-28\% & \cellcolor{green!25}0.070 & \cellcolor{green!25}250\% & \cellcolor{red!25}0.391 & \cellcolor{red!25}-19\% & \cellcolor{yellow!25}0.079 & \cellcolor{yellow!25}NaN & \cellcolor{red!25}0.202 & \cellcolor{red!25}-11\% \\
    ROS   & \cellcolor{red!25}0.465 & \cellcolor{red!25}-5\% & \cellcolor{green!25}0.006 & \cellcolor{green!25}NaN & \cellcolor{red!25}0.355 & \cellcolor{red!25}-3\% & \cellcolor{green!25}0.034 & \cellcolor{green!25}70\% & \cellcolor{red!25}0.474 & \cellcolor{red!25}-2\% & \cellcolor{green!25}0.027 & \cellcolor{green!25}NaN & \cellcolor{green!25}0.227 & \cellcolor{green!25}0\% \\
    SM    & \cellcolor{red!25}0.441 & \cellcolor{red!25}-10\% & \cellcolor{green!25}0.004 & \cellcolor{green!25}NaN & \cellcolor{red!25}0.356 & \cellcolor{red!25}-3\% & \cellcolor{green!25}0.048 & \cellcolor{green!25}140\% & \cellcolor{red!25}0.481 & \cellcolor{red!25}0\% & \cellcolor{green!25}0.054 & \cellcolor{green!25}NaN & \cellcolor{green!25}0.231 & \cellcolor{green!25}2\% \\
    bSM   & \cellcolor{red!25}0.488 & \cellcolor{red!25}0\% & \cellcolor{green!25}0.004 & \cellcolor{green!25}NaN & \cellcolor{green!25}0.377 & \cellcolor{green!25}3\% & \cellcolor{green!25}0.049 & \cellcolor{green!25}145\% & \cellcolor{yellow!25}0.482 & \cellcolor{yellow!25}0\% & \cellcolor{green!25}0.066 & \cellcolor{green!25}NaN & \cellcolor{green!25}0.244 & \cellcolor{green!25}8\% \\
    SM-ENN & \cellcolor{yellow!25}0.543 & \cellcolor{yellow!25}11\% & \cellcolor{green!25}0.005 & \cellcolor{green!25}NaN & \cellcolor{red!25}0.212 & \cellcolor{red!25}-42\% & \cellcolor{green!25}0.045 & \cellcolor{green!25}125\% & \cellcolor{red!25}0.469 & \cellcolor{red!25}-3\% & \cellcolor{green!25}0.069 & \cellcolor{green!25}NaN & \cellcolor{red!25}0.224 & \cellcolor{red!25}-1\% \\
    SM-TL & \cellcolor{green!25}0.521 & \cellcolor{green!25}6\% & \cellcolor{green!25}0.005 & \cellcolor{green!25}NaN & \cellcolor{red!25}0.324 & \cellcolor{red!25}-11\% & \cellcolor{green!25}0.040 & \cellcolor{green!25}100\% & \cellcolor{red!25}0.481 & \cellcolor{red!25}0\% & \cellcolor{green!25}0.054 & \cellcolor{green!25}NaN & \cellcolor{green!25}0.238 & \cellcolor{green!25}5\% \\
    IHT   & \cellcolor{red!25}0.158 & \cellcolor{red!25}-68\% & \cellcolor{green!25}0.015 & \cellcolor{green!25}NaN & \cellcolor{red!25}0.246 & \cellcolor{red!25}-33\% & \cellcolor{yellow!25}0.085 & \cellcolor{yellow!25}325\% & \cellcolor{red!25}0.280 & \cellcolor{red!25}-42\% & \cellcolor{green!25}0.070 & \cellcolor{green!25}NaN & \cellcolor{red!25}0.142 & \cellcolor{red!25}-37\% \\
    RanE  & \cellcolor{green!25}0.505 & \cellcolor{green!25}3\% & \cellcolor{red!25}0.000 & \cellcolor{red!25}NaN & \cellcolor{yellow!25}0.385 & \cellcolor{yellow!25}5\% & \cellcolor{red!25}0.016 & \cellcolor{red!25}-20\% & \cellcolor{red!25}0.481 & \cellcolor{red!25}0\% & \cellcolor{green!25}0.004 & \cellcolor{green!25}NaN & \cellcolor{green!25}0.232 & \cellcolor{green!25}3\% \\
    SemE  & \cellcolor{green!25}0.505 & \cellcolor{green!25}3\% & \cellcolor{green!25}0.028 & \cellcolor{green!25}NaN & \cellcolor{yellow!25}0.385 & \cellcolor{yellow!25}5\% & \cellcolor{red!25}0.014 & \cellcolor{red!25}-30\% & \cellcolor{red!25}0.481 & \cellcolor{red!25}0\% & \cellcolor{green!25}0.027 & \cellcolor{green!25}NaN & \cellcolor{green!25}0.240 & \cellcolor{green!25}6\% \\
    SupE  & \cellcolor{green!25}0.505 & \cellcolor{green!25}3\% & \cellcolor{yellow!25}0.198 & \cellcolor{yellow!25}NaN & \cellcolor{yellow!25}0.385 & \cellcolor{yellow!25}5\% & \cellcolor{green!25}0.037 & \cellcolor{green!25}85\% & \cellcolor{red!25}0.481 & \cellcolor{red!25}0\% & \cellcolor{red!25}0.000 & \cellcolor{red!25}NaN & \cellcolor{yellow!25}0.268 & \cellcolor{yellow!25}19\% \\
    \bottomrule
    \end{tabular}%
  \label{tab:addlabel}%
\end{table}%

% Table generated by Excel2LaTeX from sheet 'by class to art'
\begin{table}[H]
  \centering
  \caption{The comparison of RanE, SemE, and SupE with the state-of-the-art methods for multi-class
classification on Irvine Messages dataset. The classification executed with the Random Forest classifier and the F-measure averaged over all classes as the quality measure.}
  \setlength\tabcolsep{4pt}
    \begin{tabular}{l|rrrrrrrrrrrr|MO}
    \multicolumn{15}{c}{\textbf{Irvine Messages}} \\
    \midrule
    \multicolumn{1}{r}{} & \multicolumn{12}{c|}{Class label}                                                             & \multicolumn{1}{r}{\multirow{4}{0.07\linewidth}{Avg [F-me asure]}} & \multicolumn{1}{c}{\multirow{4}{0.04\linewidth}{Gain [\%]}} \\
    \multicolumn{1}{r}{} & \multicolumn{2}{c}{Continuing} & \multicolumn{2}{c}{Dissolving} & \multicolumn{2}{c}{Growing} & \multicolumn{2}{c}{Merging} & \multicolumn{2}{c}{Shrinking} & \multicolumn{2}{c|}{Splitting} &       &  \\
\cmidrule{1-13}    Count & \multicolumn{2}{c}{509} & \multicolumn{2}{c}{811} & \multicolumn{2}{c}{681} & \multicolumn{2}{c}{81} & \multicolumn{2}{c}{750} & \multicolumn{2}{c|}{36} &       &  \\
    IR    & \multicolumn{2}{c}{1.6} & \multicolumn{2}{c}{1.0} & \multicolumn{2}{c}{1.2} & \multicolumn{2}{c}{10.0} & \multicolumn{2}{c}{1.1} & \multicolumn{2}{c|}{22.5} &       &  \\
    \midrule
    None  & 0.184 & \multicolumn{1}{c}{-} & 0.230 & \multicolumn{1}{c}{-} & 0.267 & \multicolumn{1}{c}{-} & 0.061 & \multicolumn{1}{c}{-} & 0.532 & \multicolumn{1}{c}{-} & 0.000 & \multicolumn{1}{c|}{-} & 0.212 & \multicolumn{1}{c}{-} \\
    RUS   & \cellcolor{green!25}0.217 & \cellcolor{green!25}18\% & \cellcolor{red!25}0.176 & \cellcolor{red!25}-23\% & \cellcolor{red!25}0.196 & \cellcolor{red!25}-27\% & \cellcolor{green!25}0.077 & \cellcolor{green!25}26\% & \cellcolor{red!25}0.338 & \cellcolor{red!25}-36\% & \cellcolor{green!25}0.112 & \cellcolor{green!25}NaN & \cellcolor{red!25}0.186 & \cellcolor{red!25}-12\% \\
    ROS   & \cellcolor{green!25}0.198 & \cellcolor{green!25}8\% & \cellcolor{red!25}0.213 & \cellcolor{red!25}-7\% & \cellcolor{green!25}0.281 & \cellcolor{green!25}5\% & \cellcolor{green!25}0.094 & \cellcolor{green!25}54\% & \cellcolor{green!25}0.530 & \cellcolor{green!25}0\% & \cellcolor{green!25}0.026 & \cellcolor{green!25}NaN & \cellcolor{green!25}0.224 & \cellcolor{green!25}6\% \\
    SM    & \cellcolor{green!25}0.197 & \cellcolor{green!25}7\% & \cellcolor{red!25}0.224 & \cellcolor{red!25}-3\% & \cellcolor{green!25}0.294 & \cellcolor{green!25}10\% & \cellcolor{green!25}0.094 & \cellcolor{green!25}54\% & \cellcolor{red!25}0.522 & \cellcolor{red!25}-2\% & \cellcolor{green!25}0.072 & \cellcolor{green!25}NaN & \cellcolor{green!25}0.234 & \cellcolor{green!25}10\% \\
    bSM   & \cellcolor{green!25}0.192 & \cellcolor{green!25}4\% & \cellcolor{red!25}0.226 & \cellcolor{red!25}-2\% & \cellcolor{green!25}0.285 & \cellcolor{green!25}7\% & \cellcolor{green!25}0.089 & \cellcolor{green!25}46\% & \cellcolor{red!25}0.514 & \cellcolor{red!25}-3\% & \cellcolor{green!25}0.069 & \cellcolor{green!25}NaN & \cellcolor{green!25}0.229 & \cellcolor{green!25}8\% \\
    SM-ENN & \cellcolor{yellow!25}0.280 & \cellcolor{yellow!25}52\% & \cellcolor{red!25}0.084 & \cellcolor{red!25}-63\% & \cellcolor{red!25}0.253 & \cellcolor{red!25}-5\% & \cellcolor{green!25}0.087 & \cellcolor{green!25}43\% & \cellcolor{red!25}0.509 & \cellcolor{red!25}-4\% & \cellcolor{green!25}0.117 & \cellcolor{green!25}NaN & \cellcolor{green!25}0.222 & \cellcolor{green!25}5\% \\
    SM-TL & \cellcolor{green!25}0.217 & \cellcolor{green!25}18\% & \cellcolor{red!25}0.179 & \cellcolor{red!25}-22\% & \cellcolor{green!25}0.280 & \cellcolor{green!25}5\% & \cellcolor{green!25}0.076 & \cellcolor{green!25}25\% & \cellcolor{red!25}0.529 & \cellcolor{red!25}-1\% & \cellcolor{green!25}0.082 & \cellcolor{green!25}NaN & \cellcolor{green!25}0.227 & \cellcolor{green!25}7\% \\
    IHT   & \cellcolor{green!25}0.223 & \cellcolor{green!25}21\% & \cellcolor{red!25}0.105 & \cellcolor{red!25}-54\% & \cellcolor{red!25}0.208 & \cellcolor{red!25}-22\% & \cellcolor{green!25}0.084 & \cellcolor{green!25}38\% & \cellcolor{red!25}0.340 & \cellcolor{red!25}-36\% & \cellcolor{green!25}0.101 & \cellcolor{green!25}NaN & \cellcolor{red!25}0.177 & \cellcolor{red!25}-17\% \\
    RanE  & \cellcolor{red!25}0.171 & \cellcolor{red!25}-7\% & \cellcolor{red!25}0.230 & \cellcolor{red!25}0\% & \cellcolor{yellow!25}0.319 & \cellcolor{yellow!25}19\% & \cellcolor{red!25}0.052 & \cellcolor{red!25}-15\% & \cellcolor{red!25}0.526 & \cellcolor{red!25}-1\% & \cellcolor{green!25}0.068 & \cellcolor{green!25}NaN & \cellcolor{green!25}0.228 & \cellcolor{green!25}8\% \\
    SemE  & \cellcolor{red!25}0.179 & \cellcolor{red!25}-3\% & \cellcolor{yellow!25}0.256 & \cellcolor{yellow!25}11\% & \cellcolor{green!25}0.278 & \cellcolor{green!25}4\% & \cellcolor{green!25}0.087 & \cellcolor{green!25}43\% & \cellcolor{yellow!25}0.546 & \cellcolor{yellow!25}3\% & \cellcolor{yellow!25}0.141 & \cellcolor{yellow!25}NaN & \cellcolor{green!25}0.248 & \cellcolor{green!25}17\% \\
    SupE  & \cellcolor{red!25}0.196 & \cellcolor{red!25}7\% & \cellcolor{green!25}0.249 & \cellcolor{green!25}8\% & \cellcolor{green!25}0.292 & \cellcolor{green!25}9\% & \cellcolor{yellow!25}0.174 & \cellcolor{yellow!25}185\% & \cellcolor{yellow!25}0.546 & \cellcolor{yellow!25}3\% & \cellcolor{green!25}0.094 & \cellcolor{green!25}NaN & \cellcolor{yellow!25}0.259 & \cellcolor{yellow!25}22\% \\
    \bottomrule
    \end{tabular}%
  \label{tab:addlabel}%
\end{table}%

% Table generated by Excel2LaTeX from sheet 'by class to art'
\begin{table}[H]
  \centering
  \caption{The comparison of RanE, SemE, and SupE with the state-of-the-art methods for multi-class
classification on Loans dataset. The classification executed with the Random Forest classifier and the F-measure averaged over all classes as the quality measure.}
  \setlength\tabcolsep{4pt}
    \begin{tabular}{l|rrrrrrrrrrrr|MO}
    \multicolumn{15}{c}{\textbf{Loans}} \\
    \midrule
    \multicolumn{1}{r}{} & \multicolumn{12}{c|}{Class label}                                                             & \multicolumn{1}{r}{\multirow{4}{0.07\linewidth}{Avg [F-me asure]}} & \multicolumn{1}{c}{\multirow{4}{0.04\linewidth}{Gain [\%]}} \\
    \multicolumn{1}{r}{} & \multicolumn{2}{c}{Continuing} & \multicolumn{2}{c}{Dissolving} & \multicolumn{2}{c}{Growing} & \multicolumn{2}{c}{Merging} & \multicolumn{2}{c}{Shrinking} & \multicolumn{2}{c|}{Splitting} &       &  \\
\cmidrule{1-13}    Count & \multicolumn{2}{c}{805} & \multicolumn{2}{c}{7682} & \multicolumn{2}{c}{2913} & \multicolumn{2}{c}{46} & \multicolumn{2}{c}{2467} & \multicolumn{2}{c|}{29} &       &  \\
    IR    & \multicolumn{2}{c}{9.5} & \multicolumn{2}{c}{1.0} & \multicolumn{2}{c}{2.6} & \multicolumn{2}{c}{167.0} & \multicolumn{2}{c}{3.1} & \multicolumn{2}{c|}{264.9} &       &  \\
    \midrule
    None  & 0.038 & \multicolumn{1}{c}{-} & 0.638 & \multicolumn{1}{c}{-} & 0.207 & \multicolumn{1}{c}{-} & 0.061 & \multicolumn{1}{c}{-} & 0.193 & \multicolumn{1}{c}{-} & 0.000 & \multicolumn{1}{c|}{-} & 0.180 & \multicolumn{1}{c}{-} \\
    RUS   & \cellcolor{green!25}0.091 & \cellcolor{green!25}139\% & \cellcolor{red!25}0.212 & \cellcolor{red!25}-67\% & \cellcolor{red!25}0.122 & \cellcolor{red!25}-41\% & \cellcolor{red!25}0.000 & \cellcolor{red!25}NaN & \cellcolor{red!25}0.114 & \cellcolor{red!25}-41\% & \cellcolor{green!25}0.034 & \cellcolor{green!25}NaN & \cellcolor{red!25}0.098 & \cellcolor{red!25}-46\% \\
    ROS   & \cellcolor{green!25}0.070 & \cellcolor{green!25}84\% & \cellcolor{red!25}0.591 & \cellcolor{red!25}-7\% & \cellcolor{green!25}0.260 & \cellcolor{green!25}26\% & \cellcolor{red!25}0.012 & \cellcolor{red!25}-80\% & \cellcolor{green!25}0.248 & \cellcolor{green!25}28\% & \cellcolor{green!25}0.021 & \cellcolor{green!25}NaN & \cellcolor{green!25}0.199 & \cellcolor{green!25}11\% \\
    SM    & \cellcolor{green!25}0.081 & \cellcolor{green!25}113\% & \cellcolor{red!25}0.532 & \cellcolor{red!25}-17\% & \cellcolor{green!25}0.260 & \cellcolor{green!25}26\% & \cellcolor{red!25}0.001 & \cellcolor{red!25}-98\% & \cellcolor{green!25}0.260 & \cellcolor{green!25}35\% & \cellcolor{green!25}0.057 & \cellcolor{green!25}NaN & \cellcolor{green!25}0.201 & \cellcolor{green!25}12\% \\
    bSM   & \cellcolor{green!25}0.086 & \cellcolor{green!25}126\% & \cellcolor{red!25}0.537 & \cellcolor{red!25}-16\% & \cellcolor{green!25}0.262 & \cellcolor{green!25}27\% & \cellcolor{red!25}0.016 & \cellcolor{red!25}-74\% & \cellcolor{green!25}0.266 & \cellcolor{green!25}38\% & \cellcolor{green!25}0.002 & \cellcolor{green!25}NaN & \cellcolor{green!25}0.193 & \cellcolor{green!25}7\% \\
    SM-ENN & \cellcolor{green!25}0.099 & \cellcolor{green!25}161\% & \cellcolor{red!25}0.375 & \cellcolor{red!25}-41\% & \cellcolor{yellow!25}0.305 & \cellcolor{yellow!25}47\% & \cellcolor{red!25}0.003 & \cellcolor{red!25}-95\% & \cellcolor{yellow!25}0.307 & \cellcolor{yellow!25}59\% & \cellcolor{green!25}0.043 & \cellcolor{green!25}NaN & \cellcolor{green!25}0.189 & \cellcolor{green!25}5\% \\
    SM-TL & \cellcolor{green!25}0.083 & \cellcolor{green!25}118\% & \cellcolor{red!25}0.543 & \cellcolor{red!25}-15\% & \cellcolor{green!25}0.262 & \cellcolor{green!25}27\% & \cellcolor{red!25}0.007 & \cellcolor{red!25}-89\% & \cellcolor{green!25}0.257 & \cellcolor{green!25}33\% & \cellcolor{green!25}0.042 & \cellcolor{green!25}NaN & \cellcolor{green!25}0.198 & \cellcolor{green!25}10\% \\
    IHT   & \cellcolor{green!25}0.125 & \cellcolor{green!25}229\% & \cellcolor{red!25}0.291 & \cellcolor{red!25}-54\% & \cellcolor{green!25}0.203 & \cellcolor{green!25}-2\% & \cellcolor{red!25}0.001 & \cellcolor{red!25}-98\% & \cellcolor{green!25}0.263 & \cellcolor{green!25}36\% & \cellcolor{green!25}0.035 & \cellcolor{green!25}NaN & \cellcolor{red!25}0.155 & \cellcolor{red!25}-14\% \\
    RanE  & \cellcolor{green!25}0.042 & \cellcolor{green!25}11\% & \cellcolor{green!25}0.678 & \cellcolor{green!25}6\% & \cellcolor{green!25}0.222 & \cellcolor{green!25}7\% & \cellcolor{red!25}0.012 & \cellcolor{red!25}-80\% & \cellcolor{red!25}0.189 & \cellcolor{red!25}-2\% & \cellcolor{red!25}0.000 & \cellcolor{red!25}0\% & \cellcolor{green!25}0.189 & \cellcolor{green!25}5\% \\
    SemE  & \cellcolor{green!25}0.077 & \cellcolor{green!25}103\% & \cellcolor{yellow!25}0.689 & \cellcolor{yellow!25}8\% & \cellcolor{green!25}0.243 & \cellcolor{green!25}17\% & \cellcolor{red!25}0.000 & \cellcolor{red!25}NaN & \cellcolor{red!25}0.181 & \cellcolor{red!25}-6\% & \cellcolor{green!25}0.093 & \cellcolor{green!25}NaN & \cellcolor{green!25}0.214 & \cellcolor{green!25}19\% \\
    SupE  & \cellcolor{yellow!25}0.128 & \cellcolor{yellow!25}237\% & \cellcolor{yellow!25}0.689 & \cellcolor{yellow!25}8\% & \cellcolor{green!25}0.258 & \cellcolor{green!25}25\% & \cellcolor{red!25}0.000 & \cellcolor{red!25}NaN & \cellcolor{red!25}0.189 & \cellcolor{red!25}-2\% & \cellcolor{yellow!25}0.116 & \cellcolor{yellow!25}NaN & \cellcolor{yellow!25}0.230 & \cellcolor{yellow!25}28\% \\
    \bottomrule
    \end{tabular}%
  \label{tab:addlabel}%
\end{table}%

% Table generated by Excel2LaTeX from sheet 'by class to art'
\begin{table}[H]
  \centering
  \caption{The comparison of RanE, SemE, and SupE with the state-of-the-art methods for multi-class
classification on Facebook dataset. The classification executed with the Random Forest classifier and the F-measure averaged over all classes as the quality measure.}
  \setlength\tabcolsep{4pt}
    \begin{tabular}{l|rrrrrrrrrrrr|MO}
    \multicolumn{15}{c}{\textbf{Facebook}} \\
    \midrule
    \multicolumn{1}{r}{} & \multicolumn{12}{c|}{Class label}                                                             & \multicolumn{1}{r}{\multirow{4}{0.07\linewidth}{Avg [F-me asure]}} & \multicolumn{1}{c}{\multirow{4}{0.04\linewidth}{Gain [\%]}} \\
    \multicolumn{1}{r}{} & \multicolumn{2}{c}{Continuing} & \multicolumn{2}{c}{Dissolving} & \multicolumn{2}{c}{Growing} & \multicolumn{2}{c}{Merging} & \multicolumn{2}{c}{Shrinking} & \multicolumn{2}{c|}{Splitting} &       &  \\
\cmidrule{1-13}    Count & \multicolumn{2}{c}{2296} & \multicolumn{2}{c}{3178} & \multicolumn{2}{c}{4338} & \multicolumn{2}{c}{690} & \multicolumn{2}{c}{4282} & \multicolumn{2}{c|}{333} &       &  \\
    IR    & \multicolumn{2}{c}{1.9} & \multicolumn{2}{c}{1.4} & \multicolumn{2}{c}{1.0} & \multicolumn{2}{c}{6.3} & \multicolumn{2}{c}{1.0} & \multicolumn{2}{c|}{13.0} &       &  \\
    \midrule
    None  & 0.196 & \multicolumn{1}{c}{-} & 0.349 & \multicolumn{1}{c}{-} & 0.393 & \multicolumn{1}{c}{-} & 0.013 & \multicolumn{1}{c}{-} & 0.585 & \multicolumn{1}{c}{-} & 0.023 & \multicolumn{1}{c|}{-} & 0.260 & \multicolumn{1}{c}{-} \\
    RUS   & \cellcolor{red!25}0.121 & \cellcolor{red!25}-38\% & \cellcolor{red!25}0.280 & \cellcolor{red!25}-20\% & \cellcolor{red!25}0.227 & \cellcolor{red!25}-42\% & \cellcolor{green!25}0.089 & \cellcolor{green!25}585\% & \cellcolor{red!25}0.308 & \cellcolor{red!25}-47\% & \cellcolor{green!25}0.092 & \cellcolor{green!25}300\% & \cellcolor{red!25}0.186 & \cellcolor{red!25}-28\% \\
    ROS   & \cellcolor{green!25}0.223 & \cellcolor{green!25}14\% & \cellcolor{red!25}0.323 & \cellcolor{red!25}-7\% & \cellcolor{red!25}0.347 & \cellcolor{red!25}-12\% & \cellcolor{green!25}0.069 & \cellcolor{green!25}431\% & \cellcolor{red!25}0.531 & \cellcolor{red!25}-9\% & \cellcolor{green!25}0.071 & \cellcolor{green!25}209\% & \cellcolor{green!25}0.261 & \cellcolor{green!25}0\% \\
    SM    & \cellcolor{green!25}0.216 & \cellcolor{green!25}10\% & \cellcolor{red!25}0.316 & \cellcolor{red!25}-9\% & \cellcolor{red!25}0.349 & \cellcolor{red!25}-11\% & \cellcolor{green!25}0.075 & \cellcolor{green!25}477\% & \cellcolor{red!25}0.520 & \cellcolor{red!25}-11\% & \cellcolor{green!25}0.116 & \cellcolor{green!25}404\% & \cellcolor{green!25}0.265 & \cellcolor{green!25}2\% \\
    bSM   & \cellcolor{green!25}0.218 & \cellcolor{green!25}11\% & \cellcolor{green!25}0.359 & \cellcolor{green!25}3\% & \cellcolor{red!25}0.363 & \cellcolor{red!25}-8\% & \cellcolor{green!25}0.080 & \cellcolor{green!25}515\% & \cellcolor{red!25}0.548 & \cellcolor{red!25}-6\% & \cellcolor{green!25}0.158 & \cellcolor{green!25}587\% & \cellcolor{green!25}0.288 & \cellcolor{green!25}11\% \\
    SM-ENN & \cellcolor{yellow!25}0.279 & \cellcolor{yellow!25}42\% & \cellcolor{red!25}0.317 & \cellcolor{red!25}-9\% & \cellcolor{red!25}0.205 & \cellcolor{red!25}-48\% & \cellcolor{yellow!25}0.103 & \cellcolor{yellow!25}692\% & \cellcolor{red!25}0.503 & \cellcolor{red!25}-14\% & \cellcolor{yellow!25}0.164 & \cellcolor{yellow!25}613\% & \cellcolor{green!25}0.262 & \cellcolor{green!25}1\% \\
    SM-TL & \cellcolor{green!25}0.249 & \cellcolor{green!25}27\% & \cellcolor{red!25}0.344 & \cellcolor{red!25}-1\% & \cellcolor{red!25}0.340 & \cellcolor{red!25}-13\% & \cellcolor{green!25}0.081 & \cellcolor{green!25}523\% & \cellcolor{red!25}0.553 & \cellcolor{red!25}-5\% & \cellcolor{green!25}0.145 & \cellcolor{green!25}530\% & \cellcolor{green!25}0.286 & \cellcolor{green!25}10\% \\
    IHT   & \cellcolor{green!25}0.265 & \cellcolor{green!25}35\% & \cellcolor{red!25}0.287 & \cellcolor{red!25}-18\% & \cellcolor{red!25}0.257 & \cellcolor{red!25}-35\% & \cellcolor{green!25}0.108 & \cellcolor{green!25}731\% & \cellcolor{red!25}0.368 & \cellcolor{red!25}-37\% & \cellcolor{green!25}0.102 & \cellcolor{green!25}343\% & \cellcolor{red!25}0.231 & \cellcolor{red!25}-11\% \\
    RanE  & \cellcolor{green!25}0.200 & \cellcolor{green!25}2\% & \cellcolor{green!25}0.358 & \cellcolor{green!25}3\% & \cellcolor{green!25}0.402 & \cellcolor{green!25}2\% & \cellcolor{green!25}0.036 & \cellcolor{green!25}177\% & \cellcolor{yellow!25}0.609 & \cellcolor{yellow!25}4\% & \cellcolor{red!25}0.018 & \cellcolor{red!25}-22\% & \cellcolor{green!25}0.271 & \cellcolor{green!25}4\% \\
    SemE  & \cellcolor{green!25}0.215 & \cellcolor{green!25}10\% & \cellcolor{yellow!25}0.366 & \cellcolor{yellow!25}5\% & \cellcolor{green!25}0.402 & \cellcolor{green!25}2\% & \cellcolor{green!25}0.055 & \cellcolor{green!25}323\% & \cellcolor{yellow!25}0.609 & \cellcolor{yellow!25}4\% & \cellcolor{red!25}0.016 & \cellcolor{red!25}-30\% & \cellcolor{green!25}0.277 & \cellcolor{green!25}7\% \\
    SupE  & \cellcolor{green!25}0.243 & \cellcolor{green!25}24\% & \cellcolor{green!25}0.363 & \cellcolor{green!25}4\% & \cellcolor{yellow!25}0.404 & \cellcolor{yellow!25}3\% & \cellcolor{green!25}0.097 & \cellcolor{green!25}646\% & \cellcolor{green!25}0.607 & \cellcolor{green!25}4\% & \cellcolor{green!25}0.050 & \cellcolor{green!25}117\% & \cellcolor{yellow!25}0.294 & \cellcolor{yellow!25}13\% \\
    \bottomrule
    \end{tabular}%
  \label{tab:addlabel}%
\end{table}%

% Table generated by Excel2LaTeX from sheet 'by class to art'
\begin{table}[H]
  \centering
  \caption{The comparison of RanE, SemE, and SupE with the state-of-the-art methods for multi-class
classification on MIT dataset. The classification executed with the Random Forest classifier and the F-measure averaged over all classes as the quality measure.}
  \setlength\tabcolsep{4pt}
    \begin{tabular}{l|rrrrrrrrrrrr|MO}
    \multicolumn{15}{c}{\textbf{MIT}} \\
    \midrule
    \multicolumn{1}{r}{} & \multicolumn{12}{c|}{Class label}                                                             & \multicolumn{1}{r}{\multirow{4}{0.07\linewidth}{Avg [F-me asure]}} & \multicolumn{1}{c}{\multirow{4}{0.04\linewidth}{Gain [\%]}} \\
    \multicolumn{1}{r}{} & \multicolumn{2}{c}{Continuing} & \multicolumn{2}{c}{Dissolving} & \multicolumn{2}{c}{Growing} & \multicolumn{2}{c}{Merging} & \multicolumn{2}{c}{Shrinking} & \multicolumn{2}{c|}{Splitting} &       &  \\
\cmidrule{1-13}    Count & \multicolumn{2}{c}{84} & \multicolumn{2}{c}{27} & \multicolumn{2}{c}{100} & \multicolumn{2}{c}{34} & \multicolumn{2}{c}{103} & \multicolumn{2}{c|}{15} &       &  \\
    IR    & \multicolumn{2}{c}{1.2} & \multicolumn{2}{c}{3.8} & \multicolumn{2}{c}{1.0} & \multicolumn{2}{c}{3.0} & \multicolumn{2}{c}{1.0} & \multicolumn{2}{c|}{6.9} &       &  \\
    \midrule
    None  & 0.292 & \multicolumn{1}{c}{-} & 0.038 & \multicolumn{1}{c}{-} & 0.291 & \multicolumn{1}{c}{-} & 0.154 & \multicolumn{1}{c}{-} & 0.459 & \multicolumn{1}{c}{-} & 0.000 & \multicolumn{1}{c|}{-} & 0.206 & \multicolumn{1}{c}{-} \\
    RUS   & \cellcolor{red!25}0.168 & \cellcolor{red!25}-42\% & \cellcolor{green!25}0.120 & \cellcolor{green!25}216\% & \cellcolor{red!25}0.130 & \cellcolor{red!25}-55\% & \cellcolor{green!25}0.156 & \cellcolor{green!25}1\% & \cellcolor{red!25}0.146 & \cellcolor{red!25}-68\% & \cellcolor{green!25}0.069 & \cellcolor{green!25}NaN & \cellcolor{red!25}0.132 & \cellcolor{red!25}-36\% \\
    ROS   & \cellcolor{red!25}0.258 & \cellcolor{red!25}-12\% & \cellcolor{green!25}0.159 & \cellcolor{green!25}318\% & \cellcolor{red!25}0.269 & \cellcolor{red!25}-8\% & \cellcolor{red!25}0.133 & \cellcolor{red!25}-14\% & \cellcolor{red!25}0.360 & \cellcolor{red!25}-22\% & \cellcolor{green!25}0.072 & \cellcolor{green!25}NaN & \cellcolor{green!25}0.209 & \cellcolor{green!25}1\% \\
    SM    & \cellcolor{red!25}0.275 & \cellcolor{red!25}-6\% & \cellcolor{green!25}0.215 & \cellcolor{green!25}466\% & \cellcolor{red!25}0.244 & \cellcolor{red!25}-16\% & \cellcolor{green!25}0.177 & \cellcolor{green!25}15\% & \cellcolor{red!25}0.356 & \cellcolor{red!25}-22\% & \cellcolor{green!25}0.088 & \cellcolor{green!25}NaN & \cellcolor{green!25}0.226 & \cellcolor{green!25}10\% \\
    bSM   & \cellcolor{red!25}0.241 & \cellcolor{red!25}-17\% & \cellcolor{green!25}0.382 & \cellcolor{green!25}905\% & \cellcolor{red!25}0.268 & \cellcolor{red!25}-8\% & \cellcolor{green!25}0.208 & \cellcolor{green!25}35\% & \cellcolor{red!25}0.207 & \cellcolor{red!25}-55\% & \cellcolor{yellow!25}0.170 & \cellcolor{yellow!25}NaN & \cellcolor{green!25}0.246 & \cellcolor{green!25}19\% \\
    SM-ENN & \cellcolor{yellow!25}0.313 & \cellcolor{yellow!25}7\% & \cellcolor{green!25}0.226 & \cellcolor{green!25}495\% & \cellcolor{red!25}0.084 & \cellcolor{red!25}-71\% & \cellcolor{red!25}0.126 & \cellcolor{red!25}-18\% & \cellcolor{red!25}0.125 & \cellcolor{red!25}-73\% & \cellcolor{green!25}0.130 & \cellcolor{green!25}NaN & \cellcolor{red!25}0.167 & \cellcolor{red!25}-19\% \\
    SM-TL & \cellcolor{red!25}0.278 & \cellcolor{red!25}-5\% & \cellcolor{green!25}0.197 & \cellcolor{green!25}418\% & \cellcolor{red!25}0.199 & \cellcolor{red!25}-32\% & \cellcolor{red!25}0.140 & \cellcolor{red!25}-9\% & \cellcolor{red!25}0.291 & \cellcolor{red!25}-37\% & \cellcolor{green!25}0.099 & \cellcolor{green!25}NaN & \cellcolor{red!25}0.201 & \cellcolor{red!25}-2\% \\
    IHT   & \cellcolor{red!25}0.263 & \cellcolor{red!25}-10\% & \cellcolor{green!25}0.191 & \cellcolor{green!25}403\% & \cellcolor{red!25}0.236 & \cellcolor{red!25}-19\% & \cellcolor{red!25}0.131 & \cellcolor{red!25}-15\% & \cellcolor{red!25}0.292 & \cellcolor{red!25}-36\% & \cellcolor{green!25}0.121 & \cellcolor{green!25}NaN & \cellcolor{red!25}0.206 & \cellcolor{red!25}0\% \\
    RanE  & \cellcolor{green!25}0.300 & \cellcolor{green!25}3\% & \cellcolor{green!25}0.139 & \cellcolor{green!25}266\% & \cellcolor{red!25}0.264 & \cellcolor{red!25}-9\% & \cellcolor{red!25}0.134 & \cellcolor{red!25}-13\% & \cellcolor{green!25}0.488 & \cellcolor{green!25}6\% & \cellcolor{green!25}0.074 & \cellcolor{green!25}NaN & \cellcolor{green!25}0.233 & \cellcolor{green!25}13\% \\
    SemE  & \cellcolor{green!25}0.308 & \cellcolor{green!25}5\% & \cellcolor{green!25}0.245 & \cellcolor{green!25}545\% & \cellcolor{yellow!25}0.352 & \cellcolor{yellow!25}21\% & \cellcolor{green!25}0.156 & \cellcolor{green!25}1\% & \cellcolor{yellow!25}0.498 & \cellcolor{yellow!25}8\% & \cellcolor{green!25}0.167 & \cellcolor{green!25}NaN & \cellcolor{green!25}0.288 & \cellcolor{green!25}40\% \\
    SupE  & \cellcolor{red!25}0.292 & \cellcolor{red!25}0\% & \cellcolor{yellow!25}0.391 & \cellcolor{yellow!25}929\% & \cellcolor{green!25}0.305 & \cellcolor{green!25}5\% & \cellcolor{yellow!25}0.234 & \cellcolor{yellow!25}52\% & \cellcolor{red!25}0.419 & \cellcolor{red!25}-9\% & \cellcolor{green!25}0.167 & \cellcolor{green!25}NaN & \cellcolor{yellow!25}0.301 & \cellcolor{yellow!25}46\% \\
    \bottomrule
    \end{tabular}%
  \label{tab:addlabel}%
\end{table}%

% Table generated by Excel2LaTeX from sheet 'by class to art'
\begin{table}[H]
  \centering
  \caption{The comparison of RanE, SemE, and SupE with the state-of-the-art methods for multi-class
classification on Slashdot dataset. The classification executed with the Random Forest classifier and the F-measure averaged over all classes as the quality measure.}
  \setlength\tabcolsep{4pt}
    \begin{tabular}{l|rrrrrrrrrrrr|MO}
    \multicolumn{15}{c}{\textbf{Slashdot}} \\
    \midrule
    \multicolumn{1}{r}{} & \multicolumn{12}{c|}{Class label}                                                             & \multicolumn{1}{r}{\multirow{4}{0.07\linewidth}{Avg [F-me asure]}} & \multicolumn{1}{c}{\multirow{4}{0.04\linewidth}{Gain [\%]}} \\
    \multicolumn{1}{r}{} & \multicolumn{2}{c}{Continuing} & \multicolumn{2}{c}{Dissolving} & \multicolumn{2}{c}{Growing} & \multicolumn{2}{c}{Merging} & \multicolumn{2}{c}{Shrinking} & \multicolumn{2}{c|}{Splitting} &       &  \\
\cmidrule{1-13}    Count & \multicolumn{2}{c}{15631} & \multicolumn{2}{c}{76} & \multicolumn{2}{c}{5048} & \multicolumn{2}{c}{954} & \multicolumn{2}{c}{4201} & \multicolumn{2}{c|}{210} &       &  \\
    IR    & \multicolumn{2}{c}{1.0} & \multicolumn{2}{c}{205.7} & \multicolumn{2}{c}{3.1} & \multicolumn{2}{c}{16.4} & \multicolumn{2}{c}{3.7} & \multicolumn{2}{c|}{74.4} &       &  \\
    \midrule
    None  & 0.671 & \multicolumn{1}{c}{-} & 0.013 & \multicolumn{1}{c}{-} & 0.253 & \multicolumn{1}{c}{-} & 0.064 & \multicolumn{1}{c}{-} & 0.335 & \multicolumn{1}{c}{-} & 0.158 & \multicolumn{1}{c|}{-} & 0.249 & \multicolumn{1}{c}{-} \\
    RUS   & \cellcolor{red!25}0.511 & \cellcolor{red!25}-24\% & \cellcolor{green!25}0.029 & \cellcolor{green!25}123\% & \cellcolor{red!25}0.218 & \cellcolor{red!25}-14\% & \cellcolor{green!25}0.101 & \cellcolor{green!25}58\% & \cellcolor{red!25}0.293 & \cellcolor{red!25}-13\% & \cellcolor{red!25}0.088 & \cellcolor{red!25}-44\% & \cellcolor{red!25}0.207 & \cellcolor{red!25}-17\% \\
    ROS   & \cellcolor{green!25}0.712 & \cellcolor{green!25}6\% & \cellcolor{green!25}0.022 & \cellcolor{green!25}69\% & \cellcolor{red!25}0.248 & \cellcolor{red!25}-2\% & \cellcolor{red!25}0.062 & \cellcolor{red!25}-3\% & \cellcolor{green!25}0.359 & \cellcolor{green!25}7\% & \cellcolor{green!25}0.204 & \cellcolor{green!25}29\% & \cellcolor{green!25}0.268 & \cellcolor{green!25}8\% \\
    SM    & \cellcolor{green!25}0.719 & \cellcolor{green!25}7\% & \cellcolor{green!25}0.015 & \cellcolor{green!25}15\% & \cellcolor{red!25}0.249 & \cellcolor{red!25}-2\% & \cellcolor{green!25}0.079 & \cellcolor{green!25}23\% & \cellcolor{green!25}0.374 & \cellcolor{green!25}12\% & \cellcolor{green!25}0.189 & \cellcolor{green!25}20\% & \cellcolor{green!25}0.271 & \cellcolor{green!25}9\% \\
    bSM   & \cellcolor{green!25}0.722 & \cellcolor{green!25}8\% & \cellcolor{green!25}0.014 & \cellcolor{green!25}8\% & \cellcolor{red!25}0.238 & \cellcolor{red!25}-6\% & \cellcolor{green!25}0.110 & \cellcolor{green!25}72\% & \cellcolor{yellow!25}0.398 & \cellcolor{yellow!25}19\% & \cellcolor{green!25}0.187 & \cellcolor{green!25}18\% & \cellcolor{yellow!25}0.278 & \cellcolor{yellow!25}12\% \\
    SM-ENN & \cellcolor{green!25}0.734 & \cellcolor{green!25}9\% & \cellcolor{green!25}0.018 & \cellcolor{green!25}38\% & \cellcolor{red!25}0.222 & \cellcolor{red!25}-12\% & \cellcolor{green!25}0.081 & \cellcolor{green!25}27\% & \cellcolor{green!25}0.384 & \cellcolor{green!25}15\% & \cellcolor{green!25}0.193 & \cellcolor{green!25}22\% & \cellcolor{green!25}0.272 & \cellcolor{green!25}9\% \\
    SM-TL & \cellcolor{green!25}0.729 & \cellcolor{green!25}9\% & \cellcolor{green!25}0.024 & \cellcolor{green!25}85\% & \cellcolor{red!25}0.235 & \cellcolor{red!25}-7\% & \cellcolor{green!25}0.083 & \cellcolor{green!25}30\% & \cellcolor{green!25}0.387 & \cellcolor{green!25}16\% & \cellcolor{green!25}0.188 & \cellcolor{green!25}19\% & \cellcolor{green!25}0.274 & \cellcolor{green!25}10\% \\
    IHT   & \cellcolor{red!25}0.480 & \cellcolor{red!25}-28\% & \cellcolor{green!25}0.016 & \cellcolor{green!25}23\% & \cellcolor{red!25}0.238 & \cellcolor{red!25}-6\% & \cellcolor{yellow!25}0.114 & \cellcolor{yellow!25}78\% & \cellcolor{red!25}0.224 & \cellcolor{red!25}-33\% & \cellcolor{red!25}0.075 & \cellcolor{red!25}-53\% & \cellcolor{red!25}0.191 & \cellcolor{red!25}-23\% \\
    RanE  & \cellcolor{yellow!25}0.741 & \cellcolor{yellow!25}10\% & \cellcolor{red!25}0.000 & \cellcolor{red!25}NaN & \cellcolor{red!25}0.207 & \cellcolor{red!25}-18\% & \cellcolor{red!25}0.033 & \cellcolor{red!25}-48\% & \cellcolor{green!25}0.336 & \cellcolor{green!25}0\% & \cellcolor{yellow!25}0.214 & \cellcolor{yellow!25}35\% & \cellcolor{green!25}0.255 & \cellcolor{green!25}2\% \\
    SemE  & \cellcolor{green!25}0.679 & \cellcolor{green!25}1\% & \cellcolor{green!25}0.038 & \cellcolor{green!25}192\% & \cellcolor{red!25}0.253 & \cellcolor{red!25}0\% & \cellcolor{green!25}0.091 & \cellcolor{green!25}42\% & \cellcolor{green!25}0.339 & \cellcolor{green!25}1\% & \cellcolor{green!25}0.194 & \cellcolor{green!25}23\% & \cellcolor{green!25}0.266 & \cellcolor{green!25}7\% \\
    SupE  & \cellcolor{green!25}0.679 & \cellcolor{green!25}1\% & \cellcolor{yellow!25}0.125 & \cellcolor{yellow!25}862\% & \cellcolor{red!25}0.248 & \cellcolor{red!25}-2\% & \cellcolor{green!25}0.105 & \cellcolor{green!25}64\% & \cellcolor{green!25}0.337 & \cellcolor{green!25}1\% & \cellcolor{green!25}0.171 & \cellcolor{green!25}8\% & \cellcolor{yellow!25}0.278 & \cellcolor{yellow!25}12\% \\
    \bottomrule
    \end{tabular}%
  \label{tab:addlabel}%
\end{table}%

% Table generated by Excel2LaTeX from sheet 'by class to art'
\begin{table}[H]
  \centering
  \caption{The comparison of RanE, SemE, and SupE with the state-of-the-art methods for multi-class
classification on Infectious dataset. The classification executed with the Random Forest classifier and the F-measure averaged over all classes as the quality measure.}
  \setlength\tabcolsep{4pt}
    \begin{tabular}{l|rrrrrrrrrrrr|MO}
    \multicolumn{15}{c}{\textbf{Infectious}} \\
    \midrule
    \multicolumn{1}{r}{} & \multicolumn{12}{c|}{Class label}                                                             & \multicolumn{1}{r}{\multirow{4}{0.07\linewidth}{Avg [F-me asure]}} & \multicolumn{1}{c}{\multirow{4}{0.04\linewidth}{Gain [\%]}} \\
    \multicolumn{1}{r}{} & \multicolumn{2}{c}{Continuing} & \multicolumn{2}{c}{Dissolving} & \multicolumn{2}{c}{Growing} & \multicolumn{2}{c}{Merging} & \multicolumn{2}{c}{Shrinking} & \multicolumn{2}{c|}{Splitting} &       &  \\
\cmidrule{1-13}    Count & \multicolumn{2}{c}{23} & \multicolumn{2}{c}{48} & \multicolumn{2}{c}{33} & \multicolumn{2}{c}{12} & \multicolumn{2}{c}{39} & \multicolumn{2}{c|}{4} &       &  \\
    IR    & \multicolumn{2}{c}{2.1} & \multicolumn{2}{c}{1.0} & \multicolumn{2}{c}{1.5} & \multicolumn{2}{c}{4.0} & \multicolumn{2}{c}{1.2} & \multicolumn{2}{c|}{12.0} &       &  \\
    \midrule
    None  & 0.375 & \multicolumn{1}{c}{-} & 0.515 & \multicolumn{1}{c}{-} & 0.140 & \multicolumn{1}{c}{-} & 0.000 & \multicolumn{1}{c}{-} & 0.467 & \multicolumn{1}{c}{-} & 0.000 & \multicolumn{1}{c|}{-} & 0.250 & \multicolumn{1}{c}{-} \\
    RUS   & \cellcolor{red!25}0.124 & \cellcolor{red!25}-67\% & \cellcolor{red!25}0.239 & \cellcolor{red!25}-54\% & \cellcolor{green!25}0.167 & \cellcolor{green!25}19\% & \cellcolor{green!25}0.123 & \cellcolor{green!25}NaN & \cellcolor{red!25}0.331 & \cellcolor{red!25}-29\% & \cellcolor{green!25}0.117 & \cellcolor{green!25}NaN & \cellcolor{red!25}0.184 & \cellcolor{red!25}-26\% \\
    ROS   & \cellcolor{green!25}0.399 & \cellcolor{green!25}6\% & \cellcolor{red!25}0.271 & \cellcolor{red!25}-47\% & \cellcolor{green!25}0.162 & \cellcolor{green!25}16\% & \cellcolor{green!25}0.087 & \cellcolor{green!25}NaN & \cellcolor{green!25}0.492 & \cellcolor{green!25}5\% & \cellcolor{green!25}0.005 & \cellcolor{green!25}NaN & \cellcolor{red!25}0.236 & \cellcolor{red!25}-6\% \\
    SM    & \cellcolor{red!25}0.358 & \cellcolor{red!25}-5\% & \cellcolor{red!25}0.313 & \cellcolor{red!25}-39\% & \cellcolor{red!25}0.133 & \cellcolor{red!25}-5\% & \cellcolor{green!25}0.091 & \cellcolor{green!25}NaN & \cellcolor{red!25}0.381 & \cellcolor{red!25}-18\% & \cellcolor{green!25}0.003 & \cellcolor{green!25}NaN & \cellcolor{red!25}0.213 & \cellcolor{red!25}-15\% \\
    bSM   & \cellcolor{green!25}0.407 & \cellcolor{green!25}9\% & \cellcolor{red!25}0.397 & \cellcolor{red!25}-23\% & \cellcolor{green!25}0.175 & \cellcolor{green!25}25\% & \cellcolor{green!25}0.070 & \cellcolor{green!25}NaN & \cellcolor{green!25}0.567 & \cellcolor{green!25}21\% & \cellcolor{red!25}0.000 & \cellcolor{red!25}0\% & \cellcolor{green!25}0.270 & \cellcolor{green!25}8\% \\
    SM-ENN & \cellcolor{red!25}0.320 & \cellcolor{red!25}-15\% & \cellcolor{red!25}0.039 & \cellcolor{red!25}-92\% & \cellcolor{red!25}0.071 & \cellcolor{red!25}-49\% & \cellcolor{green!25}0.094 & \cellcolor{green!25}NaN & \cellcolor{red!25}0.414 & \cellcolor{red!25}-11\% & \cellcolor{green!25}0.119 & \cellcolor{green!25}NaN & \cellcolor{red!25}0.176 & \cellcolor{red!25}-30\% \\
    SM-TL & \cellcolor{green!25}0.383 & \cellcolor{green!25}2\% & \cellcolor{red!25}0.319 & \cellcolor{red!25}-38\% & \cellcolor{green!25}0.150 & \cellcolor{green!25}7\% & \cellcolor{green!25}0.101 & \cellcolor{green!25}NaN & \cellcolor{green!25}0.520 & \cellcolor{green!25}11\% & \cellcolor{green!25}0.008 & \cellcolor{green!25}NaN & \cellcolor{red!25}0.247 & \cellcolor{red!25}-1\% \\
    IHT   & \cellcolor{red!25}0.291 & \cellcolor{red!25}-22\% & \cellcolor{red!25}0.301 & \cellcolor{red!25}-42\% & \cellcolor{red!25}0.136 & \cellcolor{red!25}-3\% & \cellcolor{green!25}0.052 & \cellcolor{green!25}NaN & \cellcolor{red!25}0.433 & \cellcolor{red!25}-7\% & \cellcolor{green!25}0.102 & \cellcolor{green!25}NaN & \cellcolor{red!25}0.219 & \cellcolor{red!25}-12\% \\
    RanE  & \cellcolor{yellow!25}0.542 & \cellcolor{yellow!25}45\% & \cellcolor{red!25}0.498 & \cellcolor{red!25}-3\% & \cellcolor{red!25}0.127 & \cellcolor{red!25}-9\% & \cellcolor{red!25}0.000 & \cellcolor{red!25}0\% & \cellcolor{green!25}0.588 & \cellcolor{green!25}26\% & \cellcolor{red!25}0.000 & \cellcolor{red!25}0\% & \cellcolor{green!25}0.293 & \cellcolor{green!25}17\% \\
    SemE  & \cellcolor{green!25}0.437 & \cellcolor{green!25}17\% & \cellcolor{yellow!25}0.520 & \cellcolor{yellow!25}1\% & \cellcolor{yellow!25}0.223 & \cellcolor{yellow!25}59\% & \cellcolor{green!25}0.292 & \cellcolor{green!25}NaN & \cellcolor{green!25}0.623 & \cellcolor{green!25}33\% & \cellcolor{red!25}0.000 & \cellcolor{red!25}0\% & \cellcolor{green!25}0.349 & \cellcolor{green!25}40\% \\
    SupE  & \cellcolor{green!25}0.433 & \cellcolor{green!25}15\% & \cellcolor{red!25}0.482 & \cellcolor{red!25}-6\% & \cellcolor{red!25}0.076 & \cellcolor{red!25}-46\% & \cellcolor{yellow!25}0.530 & \cellcolor{yellow!25}NaN & \cellcolor{yellow!25}0.720 & \cellcolor{yellow!25}54\% & \cellcolor{yellow!25}0.250 & \cellcolor{yellow!25}NaN & \cellcolor{yellow!25}0.415 & \cellcolor{yellow!25}66\% \\
    \bottomrule
    \end{tabular}%
  \label{tab:addlabel}%
\end{table}%

% Table generated by Excel2LaTeX from sheet 'by class to art'
\begin{table}[H]
  \centering
  \caption{The comparison of RanE, SemE, and SupE with the state-of-the-art methods for multi-class
classification on Twitter dataset. The classification executed with the Random Forest classifier and the F-measure averaged over all classes as the quality measure.}
  \setlength\tabcolsep{4pt}
    \begin{tabular}{l|rrrrrrrrrrrr|MO}
    \multicolumn{15}{c}{\textbf{Twitter}} \\
    \midrule
    \multicolumn{1}{r}{} & \multicolumn{12}{c|}{Class label}                                                             & \multicolumn{1}{r}{\multirow{4}{0.07\linewidth}{Avg [F-me asure]}} & \multicolumn{1}{c}{\multirow{4}{0.04\linewidth}{Gain [\%]}} \\
    \multicolumn{1}{r}{} & \multicolumn{2}{c}{Continuing} & \multicolumn{2}{c}{Dissolving} & \multicolumn{2}{c}{Growing} & \multicolumn{2}{c}{Merging} & \multicolumn{2}{c}{Shrinking} & \multicolumn{2}{c|}{Splitting} &       &  \\
\cmidrule{1-13}    Count & \multicolumn{2}{c}{1795} & \multicolumn{2}{c}{3112} & \multicolumn{2}{c}{2637} & \multicolumn{2}{c}{670} & \multicolumn{2}{c}{2661} & \multicolumn{2}{c|}{354} &       &  \\
    IR    & \multicolumn{2}{c}{1.7} & \multicolumn{2}{c}{1.0} & \multicolumn{2}{c}{1.2} & \multicolumn{2}{c}{4.6} & \multicolumn{2}{c}{1.2} & \multicolumn{2}{c|}{8.8} &       &  \\
    \midrule
    None  & 0.177 & \multicolumn{1}{c}{-} & 0.336 & \multicolumn{1}{c}{-} & 0.264 & \multicolumn{1}{c}{-} & 0.012 & \multicolumn{1}{c}{-} & 0.291 & \multicolumn{1}{c}{-} & 0.064 & \multicolumn{1}{c|}{-} & 0.191 & \multicolumn{1}{c}{-} \\
    RUS   & \cellcolor{green!25}0.182 & \cellcolor{green!25}3\% & \cellcolor{red!25}0.230 & \cellcolor{red!25}-32\% & \cellcolor{red!25}0.210 & \cellcolor{red!25}-20\% & \cellcolor{green!25}0.092 & \cellcolor{green!25}667\% & \cellcolor{red!25}0.137 & \cellcolor{red!25}-53\% & \cellcolor{yellow!25}0.194 & \cellcolor{yellow!25}203\% & \cellcolor{red!25}0.174 & \cellcolor{red!25}-9\% \\
    ROS   & \cellcolor{red!25}0.165 & \cellcolor{red!25}-7\% & \cellcolor{red!25}0.267 & \cellcolor{red!25}-21\% & \cellcolor{red!25}0.204 & \cellcolor{red!25}-23\% & \cellcolor{green!25}0.047 & \cellcolor{green!25}292\% & \cellcolor{red!25}0.200 & \cellcolor{red!25}-31\% & \cellcolor{green!25}0.086 & \cellcolor{green!25}34\% & \cellcolor{red!25}0.162 & \cellcolor{red!25}-15\% \\
    SM    & \cellcolor{red!25}0.158 & \cellcolor{red!25}-11\% & \cellcolor{red!25}0.222 & \cellcolor{red!25}-34\% & \cellcolor{red!25}0.182 & \cellcolor{red!25}-31\% & \cellcolor{green!25}0.053 & \cellcolor{green!25}342\% & \cellcolor{red!25}0.212 & \cellcolor{red!25}-27\% & \cellcolor{green!25}0.122 & \cellcolor{green!25}91\% & \cellcolor{red!25}0.158 & \cellcolor{red!25}-17\% \\
    bSM   & \cellcolor{green!25}0.222 & \cellcolor{green!25}25\% & \cellcolor{green!25}0.346 & \cellcolor{green!25}3\% & \cellcolor{yellow!25}0.265 & \cellcolor{yellow!25}0\% & \cellcolor{green!25}0.020 & \cellcolor{green!25}67\% & \cellcolor{red!25}0.290 & \cellcolor{red!25}0\% & \cellcolor{green!25}0.082 & \cellcolor{green!25}28\% & \cellcolor{green!25}0.204 & \cellcolor{green!25}7\% \\
    SM-ENN & \cellcolor{green!25}0.229 & \cellcolor{green!25}29\% & \cellcolor{red!25}0.266 & \cellcolor{red!25}-21\% & \cellcolor{red!25}0.219 & \cellcolor{red!25}-17\% & \cellcolor{green!25}0.075 & \cellcolor{green!25}525\% & \cellcolor{red!25}0.211 & \cellcolor{red!25}-27\% & \cellcolor{green!25}0.159 & \cellcolor{green!25}148\% & \cellcolor{green!25}0.193 & \cellcolor{green!25}1\% \\
    SM-TL & \cellcolor{yellow!25}0.258 & \cellcolor{yellow!25}46\% & \cellcolor{red!25}0.293 & \cellcolor{red!25}-13\% & \cellcolor{red!25}0.201 & \cellcolor{red!25}-24\% & \cellcolor{green!25}0.058 & \cellcolor{green!25}383\% & \cellcolor{red!25}0.215 & \cellcolor{red!25}-26\% & \cellcolor{green!25}0.126 & \cellcolor{green!25}97\% & \cellcolor{green!25}0.192 & \cellcolor{green!25}1\% \\
    IHT   & \cellcolor{green!25}0.212 & \cellcolor{green!25}20\% & \cellcolor{red!25}0.271 & \cellcolor{red!25}-19\% & \cellcolor{red!25}0.237 & \cellcolor{red!25}-10\% & \cellcolor{yellow!25}0.107 & \cellcolor{yellow!25}792\% & \cellcolor{red!25}0.216 & \cellcolor{red!25}-26\% & \cellcolor{green!25}0.175 & \cellcolor{green!25}173\% & \cellcolor{green!25}0.203 & \cellcolor{green!25}6\% \\
    RanE  & \cellcolor{green!25}0.216 & \cellcolor{green!25}22\% & \cellcolor{green!25}0.342 & \cellcolor{green!25}2\% & \cellcolor{red!25}0.241 & \cellcolor{red!25}-9\% & \cellcolor{green!25}0.054 & \cellcolor{green!25}350\% & \cellcolor{red!25}0.232 & \cellcolor{red!25}-20\% & \cellcolor{green!25}0.082 & \cellcolor{green!25}28\% & \cellcolor{green!25}0.195 & \cellcolor{green!25}2\% \\
    SemE  & \cellcolor{red!25}0.174 & \cellcolor{red!25}-2\% & \cellcolor{yellow!25}0.359 & \cellcolor{yellow!25}7\% & \cellcolor{red!25}0.264 & \cellcolor{red!25}0\% & \cellcolor{green!25}0.032 & \cellcolor{green!25}167\% & \cellcolor{yellow!25}0.306 & \cellcolor{yellow!25}5\% & \cellcolor{green!25}0.088 & \cellcolor{green!25}38\% & \cellcolor{green!25}0.204 & \cellcolor{green!25}7\% \\
    SupE  & \cellcolor{green!25}0.192 & \cellcolor{green!25}8\% & \cellcolor{green!25}0.342 & \cellcolor{green!25}2\% & \cellcolor{red!25}0.264 & \cellcolor{red!25}0\% & \cellcolor{green!25}0.083 & \cellcolor{green!25}592\% & \cellcolor{yellow!25}0.306 & \cellcolor{yellow!25}5\% & \cellcolor{green!25}0.088 & \cellcolor{green!25}38\% & \cellcolor{yellow!25}0.213 & \cellcolor{yellow!25}12\% \\
    \bottomrule
    \end{tabular}%
  \label{tab:addlabel}%
\end{table}%

\section{Enriching vs. Transfer Learning}
\label{appendix:transfer_learning}

Table~\ref{tab_comparison_tl} contains the results of the initial comparison of the Transfer Learning (TL) technique with the enriching approach. The comparison was conducted for the smallest data sets, Infectious and MIT. The model was transferred from the Twitter data set. The RanE, SemE, and SupE methods also use the Twitter data set as the external data set. The TL technique allowed us to achieve a better result than the baseline and the RanE method but was unable to surpass SemE and SupE. Nevertheless, TL performance was very satisfactory. The reason for this might be a particular data set, i.e., the Infectious and MIT data sets have critically underrepresented minority classes; hence, the classifier was unable to learn from them. Using an external data set, with more examples, turned out to be better than learning from the original data. Perhaps, applying TL to bigger data sets would not yield such spectacular results.

\begin{table}[H]
\caption{The comparison of RanE, SemE, and SupE with the Transfer Learning technique on the MIT and Infectious data sets. For all methods, the Twitter data set was used as the external data set. The classification executed with the Random Forest classifier and the F-measure averaged over all classes as the quality measure. Cells marked green indicate an improved classification quality, while cells marked red depict worse results. The best result for each data set is bolded and marked yellow.}
\label{tab_comparison_tl}
\centering
%% \tablesize{} %% You can specify the fontsize here, e.g., \tablesize{\footnotesize}. If commented out \small will be used.
\begin{tabular}{c|cc}
\toprule
\multirow{2}{1.5cm}{\centering\vfil\bfseries\textbf{Method}}    & \multicolumn{2}{c}{\bfseries Base set}\\ \cmidrule(lr){2-3}  &   \textbf{MIT}  & \textbf{Infectious}\\
\midrule
None    & 0.206	& 0.250\\
\textbf{RanE}    & \cellcolor{green!25}0.233 +13\% & \cellcolor{green!25}0.293 +17\%\\
\textbf{SemE}    & \cellcolor{green!25}0.288 +40\% & \cellcolor{green!25}0.349 +40\%\\
\textbf{SupE}    & \cellcolor{yellow!25}\textbf{0.301 +46\%} & \cellcolor{yellow!25}\textbf{0.415 +66\%}\\
\textbf{TL}  & \cellcolor{green!25}0.261 +27\% & \cellcolor{green!25}0.312 +25\%\\
\bottomrule
\end{tabular}
\end{table}

\end{document}